\documentclass[11pt]{article}
\usepackage[utf8]{inputenc}

\usepackage{acl}

\usepackage{times}
\usepackage{latexsym}
\usepackage{fontawesome5}

\usepackage[T1]{fontenc}

\usepackage{microtype}

\usepackage{inconsolata}
\usepackage{colortbl}
\usepackage{graphicx}
\usepackage{xspace}
\usepackage{xcolor}
\usepackage{amsmath}
\usepackage{amssymb}
\usepackage{booktabs} 
\usepackage{multirow}
\usepackage{amsthm}
\usepackage{bbm}
\usepackage[normalem]{ulem}
\usepackage{etoolbox}

\AtBeginEnvironment{equation}{\small}
\AtBeginEnvironment{equation*}{\small}
\AtBeginEnvironment{align}{\small}
\AtBeginEnvironment{align*}{\small}
\AtBeginEnvironment{alignat}{\small}
\AtBeginEnvironment{alignat*}{\small}
\AtBeginEnvironment{flalign}{\small}
\AtBeginEnvironment{flalign*}{\small}
\AtBeginEnvironment{gather}{\small}
\AtBeginEnvironment{gather*}{\small}
\AtBeginEnvironment{multline}{\small}
\AtBeginEnvironment{multline*}{\small}

\newcommand{\tableheader}[1]{\textbf{\boldmath #1}}

\theoremstyle{definition}
\newtheorem{theorem}{Theorem}[section]
\newtheorem{proposition}[theorem]{Proposition}
\newtheorem{lemma}[theorem]{Lemma}
\newtheorem{corollary}[theorem]{Corollary}
\newtheorem{assumption}[theorem]{Assumption}
\newtheorem{remark}[theorem]{Remark}

\newcommand{\ours}{Distributional On-Policy Distillation\xspace}
\newcommand{\shortours}{DistOPD\xspace}
\newcommand{\pref}{\pi_{\mathrm{ref}}}
\newcommand{\ptheta}{\pi_{\theta}}
\newcommand{\teacher}{\pi_{T}}
\newcommand{\KL}{\mathrm{KL}}
\newcommand{\sg}{\mathrm{sg}}
\newcommand{\E}{\mathbb{E}}
\newcommand{\R}{\mathbb{R}}
\newcommand{\hq}{\hat{q}}
\newcommand{\lopd}{Lightning OPD\xspace}

\newcommand{\Sset}{\mathcal{S}}

\usepackage{amsthm}
\usepackage{tabularx}
\newcolumntype{Y}{>{\centering\arraybackslash}X}

\title{Rethinking Offline On-Policy Distillation:\\ From Scalar Advantages to Distributional Supervision}

\author{%
  Qi Sun$^{1}$\thanks{Equal contribution.} \quad
  Siyue Zhang$^{2}$\footnotemark[1] \quad
  Tingyu Song$^3$ \quad
  Yilun Zhao$^4$ \quad
  Chen Zhao$^1$ \\
  \\
  $^1$NYU Shanghai \quad $^2$NTU \quad $^3$UCAS \quad $^4$Yale \\
  \\
  \texttt{\{qs2196, cz1285\}@nyu.edu} \quad \texttt{siyue001@e.ntu.edu.sg} \\
  \texttt{songtingyu23@mails.ucas.ac.cn} \quad \texttt{yilun.zhao@yale.edu} \\
  \\
  \vspace{-2mm} 
  \href{https://github.com/AlbertQiSun/DistOPD}{\faGithub\ \texttt{AlbertQiSun/DistOPD}}
}

\begin{document}
\maketitle

\begin{abstract}
On-policy distillation (OPD) transfers teacher capabilities through token-level feedback on student-generated trajectories, but repeatedly evaluating fresh rollouts with the teacher is computationally expensive.
Offline methods such as Lightning OPD reduce this cost by caching recorded tokens' teacher log-probabilities on fixed reference rollouts.
By extending the idea to a multi-turn agentic search setup, we find that it consistently underperforms the SFT starting point trained by the same teacher.
We empirically observe greater teacher--reference mismatch than in mathematical reasoning and post-training distributional flattening, and theoretically show that scalar advantage supervision on fixed rollouts admits solutions with flattened output distributions under suitable conditions.
Motivated by these findings, we propose \ours (\shortours), a simple yet effective offline OPD method that retains candidate-token probabilities from both reference and teacher at selected positions and aggregates the remaining parts into a residual token. 
On agentic search, \shortours{} achieves performance comparable to online OPD while improving student--teacher alignment and using training GPU-hours comparable to Lightning OPD.
\end{abstract}

\section{Introduction}
\label{sec:intro}

\begin{figure}[t]
\centering
\includegraphics[width=0.9\columnwidth]{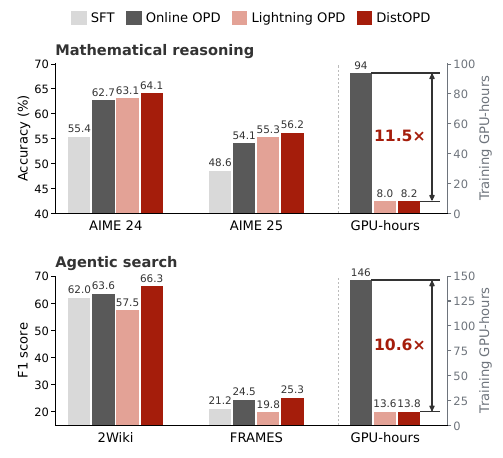}
\caption{ Lightning OPD improves mathematical reasoning efficiently but struggles on long-horizon agentic search. \textbf{Top:} On mathematical reasoning, Lightning OPD improves upon the SFT baseline with substantially fewer training GPU-hours than online OPD. \textbf{Bottom:} On agentic search, Lightning OPD instead underperforms its SFT initialization. DistOPD outperforms online OPD on the benchmarks shown while retaining the efficiency of offline training. It uses $11.5\times$ and $10.6\times$ fewer training GPU-hours than online OPD on mathematical reasoning and agentic search, respectively.}
\label{fig:teaser}
\end{figure}

On-policy distillation (OPD) transfers capabilities from a teacher model to a student through token-level supervision on student-generated trajectories \citep{agarwal2024gkd,gu2024minillm,lu2025opd}.
By obtaining teacher feedback on contexts encountered under its own policy, the student can learn to refine its own generation behavior.
However, standard OPD, which we refer to as \emph{online OPD}, requires repeated teacher evaluation of fresh student rollouts throughout training, incurring substantial computational overhead \citep{fu2026revisiting,jang2026veto,jin2026eopd,ko2026relaxed,asyncopd2026}.

Recent work, exemplified by \lopd{} \citep{wu2026lightning,wu2026lightning2}, moves teacher evaluation offline by collecting trajectories from a frozen SFT reference policy and caching teacher log-probabilities for the recorded tokens before optimization.
The student is then trained using these cached scores without further teacher inference.
On the mathematical reasoning and code generation tasks evaluated in prior work, \lopd{} achieves performance comparable to online OPD with up to a $4.0\times$ reduction in training GPU-hours \citep{wu2026lightning}.
However, it remains underexplored whether this offline supervision is effective in multi-turn agentic settings, which require the student to repeatedly incorporate external tool feedback into subsequent reasoning and actions.

To investigate this, we evaluate \lopd ~\citep{wu2026lightning} on agentic search, and results shows that across five multi-hop QA benchmarks, \lopd{} consistently underperforms its SFT initialization, whereas online OPD \citep{lu2025opd} improves performance. 
To understand why \lopd{} struggles in agentic search, our further empirical analysis into the reasons reveals larger teacher--reference mismatch \citep{shing2025taid} than in mathematical reasoning, reflected in a heavier negative tail of token-level log-probability gaps at initialization. %

Our theoretical analysis shows that, on fixed rollouts, scalar advantage regression constrains the probability of the recorded token without uniquely determining the distribution over alternatives. 
Under suitable conditions, the objective admits logit-contracted solutions without requiring full teacher alignment, consistent with the observed distributional flattening~\citep{hinton2015distilling}.
A natural remedy is to restore distributional supervision, for example through a full-vocabulary KL objective. However, caching the teacher's full distribution at every token position largely incurs storage costs, limiting the scalability of this approach to real usage.

Motivated by this, we propose \ours (\shortours), an offline OPD method that learns from compact distributional targets. 
At selected positions in fixed reference rollouts, \shortours{} caches reference and teacher probabilities over compact candidate sets, together with an aggregate tail residual token. 
The candidate sets retain high-probability tokens from both distributions and always include the recorded token.
Across five multi-hop QA benchmarks, \shortours{} outperforms its SFT initialization on every benchmark with both teachers and improves average F1 over \lopd{} by $6.0$ and $7.4$ points for the two teachers, respectively. %
Distributional diagnostics further show that \shortours improves student--teacher alignment while mitigating the distributional flattening observed with \lopd.
With the Qwen3-32B teacher, \shortours reduces training GPU-hours relative to online OPD by factors of $10.6$ and $11.5$ on agentic search and mathematical reasoning, respectively (Figure~\ref{fig:teaser}).

We highlight our contributions as follows:
\begin{itemize}
\item Through empirical and theoretical analysis, we identify limitations of \lopd{} in agentic search, where it underperforms its SFT initialization, and show that scalar advantage supervision can admit flattened output distributions.
\item We propose \ours (\shortours), which replaces scalar advantage regression with distributional supervision, mitigating flattened token probabilities and yielding more discriminative output distributions.
\item We demonstrate that \shortours{} generalizes across agentic search and mathematical reasoning, achieves performance comparable to online OPD, and substantially reduces training GPU-hours across both domains.
\end{itemize}

\section{Diagnosing Lightning OPD on Agentic Search}
\label{sec:analysis}
\label{sec:lightning}

We investigate the effectiveness of \lopd{} in agentic search through empirical evaluation and theoretical analysis of its fixed-rollout supervision.

\subsection{Preliminaries}
\label{sec:problem-setup}

OPD trains a student policy $\ptheta$ using token-level teacher feedback on its own rollouts \citep{agarwal2024gkd}.
Standard online OPD uses this signal to weight policy-gradient updates on fresh student rollouts.
To mitigate the computational cost of online OPD, recent work~\citep{wu2026lightning} explores offline OPD, which collects rollouts and caches teacher supervision before optimization, replacing repeated rollout generation and teacher evaluation with fixed trajectories \citep{wu2026lightning}.

Lightning OPD \citep{wu2026lightning, wu2026lightning2} initializes the student from an frozon SFT reference policy $\pref = \pi_{SFT}$ and caches teacher log-probabilities on fixed reference rollouts $\{x_j\sim\pref(\cdot\mid q_j)\}_{j=1}^{N}$, where $q_j$ denotes a training prompt \citep{wu2026lightning}.
For token $a_t$ at context $s_t$, it computes the advantage

\begin{equation}
A_t(\theta)=\log\teacher(a_t\mid s_t)-\log\ptheta(a_t\mid s_t),
\end{equation}

It recomputes the student term in $A_t(\theta)$ during optimization and uses the advantage-weighted update

\begin{equation}
\nabla J_{\mathrm{off}}(\theta)
=\E_{x\sim\pref}\big[\textstyle\sum_t\bar A_t(\theta)\,\nabla_\theta\log\ptheta(a_t\mid s_t)\big],
\label{eq:offline}
\end{equation}

where $\bar A_t(\theta)=\mathrm{clip}\big(\sg(A_t(\theta)),-10,10\big)$ and $\sg$ denotes stop-gradient.
The expectation is estimated over the fixed rollouts, and sums include only model-generated tokens.
Lightning OPD requires the same teacher for SFT data generation and rollout scoring, termed \emph{teacher consistency}.

We investigate the effectiveness of \lopd{} in long-horizon agentic search tasks.

\subsection{Experimental Setup}
\label{sec:exp-setup}
To assess the effectiveness and limitations of \lopd{} in agentic search, we conduct an empirical analysis comparing it with its SFT initialization and online OPD.

\paragraph{Models and Tasks.}
We distill Qwen3-4B-Base~\citep{qwen3} from two teachers: the search-specialized MiroThinker-1.7-mini~\citep{mirothinker} and the general-purpose Qwen3-32B~\citep{qwen3}.
For each teacher, we obtain $\pref$ by fine-tuning the student on that teacher's trajectories, ensuring teacher consistency~\citep{wu2026lightning}.
We train on HotpotQA~\citep{hotpotqa} and MuSiQue~\citep{musique} training questions and evaluate on five agentic search benchmarks: HotpotQA, MuSiQue, 2WikiMultiHopQA~\citep{twowikimultihopqa}, Bamboogle~\citep{bamboogle}, and FRAMES~\citep{frames}.

\paragraph{Baselines and Metrics.}
We compare \lopd{} with two baselines: SFT starting point as the frozen reference policy $\pref$ and standard online OPD.
We evaluate final answers using token-level F1 as the primary metric and exact match (EM), and report training cost in GPU-hours.
All experiments are conducted on up to two NVIDIA A100 80GB GPUs.
Implementation details and training budgets are provided in Appendix~\ref{app:impl}. %

\begin{figure}[t]
\centering
\includegraphics[width=0.9\columnwidth]{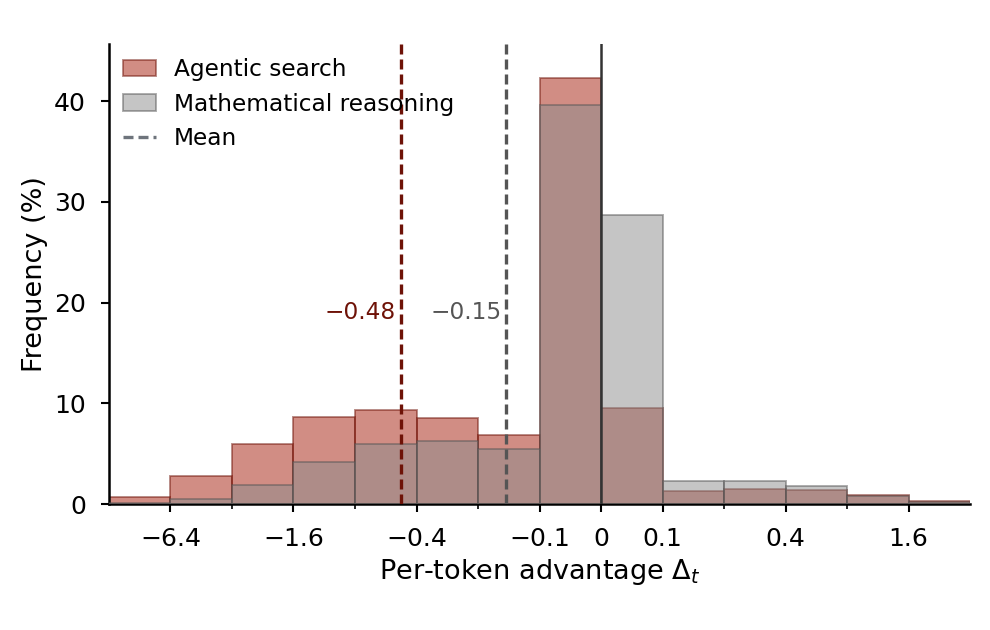}
\caption{Teacher--reference log-probability gaps $\Delta_t$ on reference-generated tokens at SFT initialization, with Qwen3-32B as the teacher. Bars show token percentages in bins whose widths increase geometrically away from zero; dashed lines indicate mean gaps. Agentic search exhibits a heavier negative tail and a more negative mean than mathematical reasoning.}
\label{fig:gap}
\end{figure}

\subsection{Empirical Findings}
\label{sec:analysis-regime}

\paragraph{Lightning OPD underperforms its SFT initialization.}
As shown in Table~\ref{tab:main}, \lopd{} underperforms $\pref$ on all five agentic search benchmarks with both teachers.
Average F1 decreases by $3.3$ points with MiroThinker-1.7-mini and $3.1$ points with Qwen3-32B, whereas online OPD improves it by $1.8$ and $3.2$ points, respectively.
To better understand this degradation, we examine changes in predictive concentration and student--teacher alignment. We use next-token entropy to measure concentration and logit margins to assess separation between competing tokens.
Relative to $\pref$, \lopd{} increases next-token entropy by $1.2$ and $1.7$ for the two teachers, respectively, and reduces logit margins to $45\%$--$70\%$ of their initial values, indicating flatter output distributions.
This flattening is accompanied by poorer teacher alignment: the measured KL divergence to the teacher increases by $0.18$ and $0.46$, respectively, whereas online OPD reduces it by $0.06$ and $0.11$.
Metric definitions and detailed measurements are provided in Appendix~\ref{app:diagnosis}.

\paragraph{Agentic search exhibits larger teacher--reference mismatch.}
We characterize teacher--reference mismatch through the token-level log-probability gap $\Delta_t$, which equals the advantage at initialization.
With the Qwen3-32B teacher, agentic search exhibits a heavier negative tail than mathematical reasoning, although both gap distributions concentrate near zero (Figure~\ref{fig:gap}).
The mean gap is $-0.48$ on agentic search, compared with $-0.15$ on mathematical reasoning; $27\%$ of generated positions have $\Delta_t<-0.4$, compared with $13\%$.
Thus, a larger fraction of recorded tokens receive negative advantages that encourage substantial reductions in their probabilities.
Our training diagnostics further show that \lopd{} produces less concentrated output distributions and increases the measured KL divergence to the teacher on agentic search (Appendix~\ref{app:diagnosis}).
These observations motivate examining whether fixed scalar supervision can adequately guide the redistribution of probability from recorded tokens to teacher-preferred alternatives.

\subsection{Theoretical Analysis}
\label{sec:analysis-theory}

We analyze the constraints imposed by scalar advantage regression on fixed reference rollouts.
The results concern state-wise objectives under independent logit parameterizations.
All proofs are deferred to Appendix~\ref{app:proofs}.

\paragraph{Scope of the Lightning OPD guarantee.}
Let $P_\theta$ and $P_{\mathrm{ref}}$ denote the joint distributions of prompts and complete interaction trajectories induced by $\ptheta$ and $\pref$, respectively.
The population online--offline update bound of \citet[Theorem~3.5]{wu2026lightning} takes the form $G\,\sigma_A\sqrt{\chi^2(P_\theta\|P_{\mathrm{ref}})}$, where $G$ bounds the token-level score-function norm and $\sigma_A$ satisfies

\begin{equation}
\E_{x\sim P_{\mathrm{ref}}}\Big[\big(\textstyle\sum_t A_t(\theta)\big)^2\Big]\le\sigma_A^2
\quad\text{for all }\theta.
\end{equation}

Under a common prompt distribution, shared environment dynamics, a finite horizon, and $P_\theta\ll P_{\mathrm{ref}}$, the trajectory-level divergence satisfies

\begin{equation}
\begin{aligned}
&\chi^2(P_\theta\|P_{\mathrm{ref}})\\
&\quad\geq
\exp\!\Big(
\E_{x\sim P_\theta}
\big[\textstyle\sum_{t=1}^{T}\KL(p_t\|\pi_t)\big]
\Big)-1.
\end{aligned}
\label{eq:chi2-growth}
\end{equation}

The sum includes only model-generated tokens; retrieved passages remain part of the conditioning contexts.
Thus, small per-token drift alone need not ensure a small trajectory-level bound.
This does not establish a large actual update discrepancy.

At initialization, the population updates coincide, but an update estimated from a fixed finite dataset can still differ due to sampling error. We therefore examine the constraints imposed by the offline objective.

\begin{assumption}[Frozen rollouts]
\label{asm:frozen}
A finite dataset $\mathcal D$ is sampled from $\pref$ and held fixed, with one recorded continuation $a_t$ per analyzed context.
Advantages are recomputed and detached during optimization.
State-wise logits are unrestricted independent variables, and teacher token probabilities are strictly positive.
\end{assumption}

\begin{theorem}[Underdetermined solutions under sampled-token supervision]
\label{thm:offline}
Under Assumption~\ref{asm:frozen}, the empirical, unclipped update corresponding to Eq.~\eqref{eq:offline} is the negative gradient of

\begin{equation}
L(\theta)
=\tfrac12\,\E_{x\sim\mathcal D}\!\Big[
\textstyle\sum_t A_t(\theta)^2\Big],
\label{eq:regression}
\end{equation}

where the expectation denotes an empirical average.
For vocabulary size $V$, the state-wise minimizers satisfy $p_t(a_t)=q_t(a_t)$ and retain $V-2$ degrees of freedom in probability space.
If $a_t$ is the unique highest-logit token and $1/V<q_t(a_t)<p_t(a_t)$, a unique contraction $\mathrm{softmax}(c z_t)$ with $c\in(0,1)$ satisfies this constraint.
Advantage clipping at $10$ replaces the squared penalty with a Huber penalty without changing these state-wise minimizers.
\end{theorem}

\begin{remark}
At initialization, $\Delta_t<0$ means that the teacher assigns a lower probability to the recorded token than the reference policy.
When the recorded token is also the unique highest-logit token and its teacher probability exceeds $1/V$, the scalar constraint can be satisfied by logit contraction, which preserves token rankings and increases entropy.
This provides a possible explanation consistent with the observed flattening, but does not establish that optimization follows this path or that contraction causes task-level degradation.
\end{remark}

We next contrast this underdetermination under fixed-rollout supervision with the constraints imposed by fresh on-policy sampling.
\begin{theorem}[Full-distribution matching under on-policy sampling]
\label{thm:online}
At a fixed context, let $p_t=\mathrm{softmax}(z_t)$ have unrestricted finite logits, and let $q_t$ be a full-support teacher distribution.
The expected unclipped update under fresh samples $a_t\sim p_t$ satisfies

\begin{equation}
\E\!\left[
\log\frac{q_t(a_t)}{p_t(a_t)}
\,\nabla_{z_t}\log p_t(a_t)
\right]
=
-\nabla_{z_t}\KL(p_t\|q_t).
\end{equation}

This update vanishes if and only if $p_t=q_t$.
\end{theorem}
Thus, under the stated assumptions, scalar advantages constrain only recorded-token probabilities in the empirical offline objective, but require full teacher alignment at stationarity of the expected online update.

\begin{remark}
The shared-fixed-point result of
\citet[Theorem~3.6]{wu2026lightning}
establishes that matching the teacher makes both updates vanish.
Our result is compatible with this guarantee: under the finite-data assumptions above, the offline update also admits additional minimizers that match the recorded-token probabilities without matching the full teacher distributions.
\end{remark}

These results show that sampled-token probability matching need not determine the full output distribution.
This motivates \ours{} (\shortours{}), an offline OPD method using cached distributional targets for richer token-level supervision.

\begin{figure*}[t]
\centering
\includegraphics[width=0.95\textwidth]{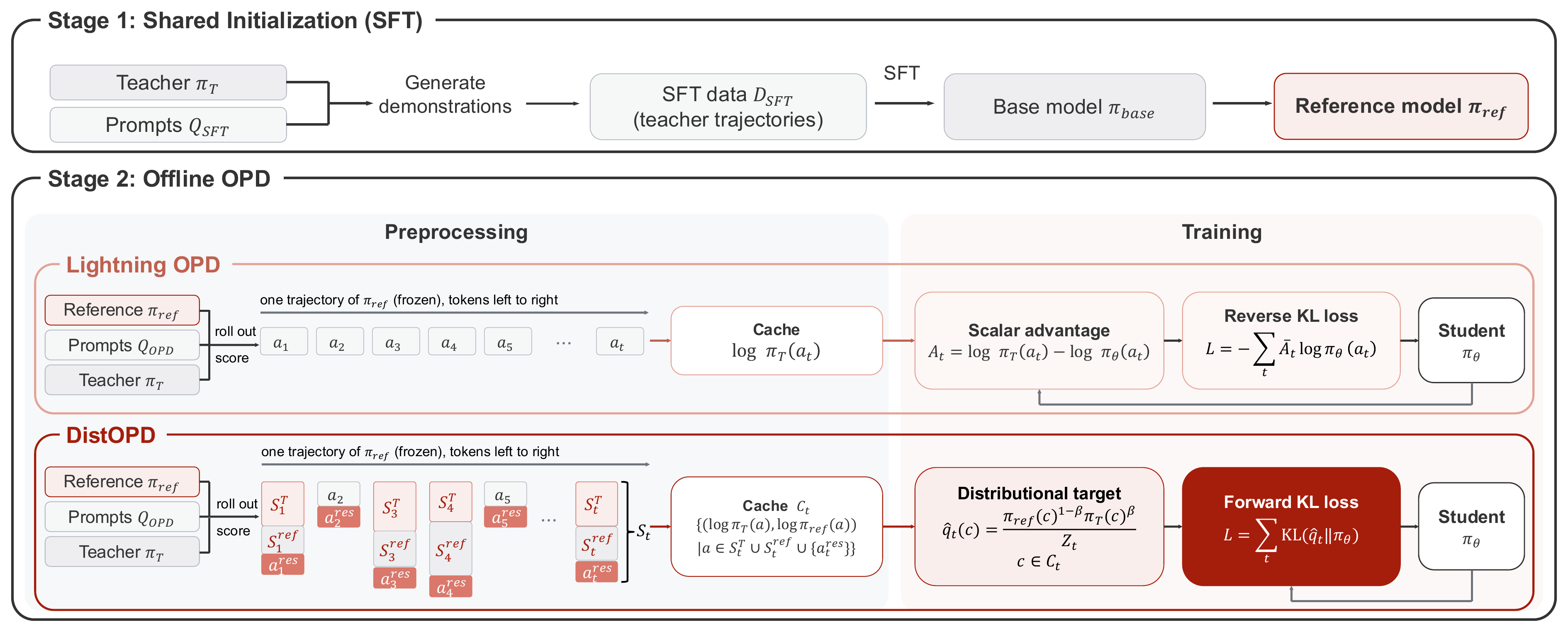}
\caption{\textbf{Comparison of Lightning OPD and DistOPD.} Both methods share SFT initialization and fixed reference rollouts. \textbf{Lightning OPD (upper row)} caches teacher log-probabilities for recorded tokens and trains with scalar advantages. \textbf{DistOPD (lower row)} caches reference and teacher distributions over selected candidates and a residual bucket, retaining only the recorded token and bucket at unselected positions. It trains with forward KL against fixed interpolated targets. Neither method requires teacher inference during training.}
\label{fig:method}
\end{figure*}

\begin{table*}[t]
\centering
\small
\setlength{\tabcolsep}{3pt}
\begin{tabularx}{0.9\textwidth}{@{}l *{13}{Y}@{}}
\toprule
\multirow{2}{*}{\tableheader{Model / Method}}
& \multicolumn{2}{c}{\tableheader{HotpotQA}}
& \multicolumn{2}{c}{\tableheader{2Wiki}}
& \multicolumn{2}{c}{\tableheader{Bamboogle}}
& \multicolumn{2}{c}{\tableheader{MuSiQue}}
& \multicolumn{2}{c}{\tableheader{FRAMES}}
& \multicolumn{2}{c}{\tableheader{Avg.}}
& \multirow{2}{*}{\tableheader{GPU-h$\downarrow$}} \\
\cmidrule(lr){2-3}\cmidrule(lr){4-5}\cmidrule(lr){6-7}\cmidrule(lr){8-9}\cmidrule(lr){10-11}\cmidrule(lr){12-13}
& \tableheader{EM$\uparrow$} & \tableheader{F1$\uparrow$}
& \tableheader{EM$\uparrow$} & \tableheader{F1$\uparrow$}
& \tableheader{EM$\uparrow$} & \tableheader{F1$\uparrow$}
& \tableheader{EM$\uparrow$} & \tableheader{F1$\uparrow$}
& \tableheader{EM$\uparrow$} & \tableheader{F1$\uparrow$}
& \tableheader{EM$\uparrow$} & \tableheader{F1$\uparrow$}
& \\
\midrule
\multicolumn{14}{@{}l}{\tableheader{\textit{Teacher models}}} \\
\quad MiroThinker-1.7-mini & 38.3 & 64.8 & 46.1 & 72.6 & 42.2 & 74.5 & 28.5 & 38.2 & 29.8 & 41.7 & 37.0 & 58.4 & -- \\
\quad Qwen3-32B & 35.2 & 60.1 & 43.5 & 69.7 & 40.8 & 73.1 & 27.4 & 33.6 & 21.2 & 36.8 & 33.6 & 54.7 & -- \\
\midrule
\multicolumn{14}{@{}l}{\tableheader{\textit{Student: Qwen3-4B-Base \enspace$\cdot$\enspace Teacher: MiroThinker-1.7-mini}}} \\
\quad SFT ($\pref$) & 31.6 & 56.2 & \underline{41.1} & 63.4 & 39.2 & 64.2 & \underline{23.3} & 30.3 & 16.2 & 24.8 & \underline{30.3} & 47.8 & -- \\
\quad Online OPD & \underline{32.8} & \underline{56.7} & 38.6 & \underline{67.3} & \underline{41.2} & \underline{65.5} & 19.6 & \textbf{32.5} & \textbf{18.3} & \underline{26.1} & 30.1 & \underline{49.6} & 168.1 \\
\quad \lopd{} & 29.2 & 51.2 & 29.5 & 61.2 & 34.8 & 62.1 & 11.0 & 26.7 & 12.9 & 21.4 & 23.5 & 44.5 & \textbf{16.7} \\
\rowcolor{gray!20} \quad \shortours{} & \textbf{34.3} & \textbf{58.8} & \textbf{41.8} & \textbf{67.4} & \textbf{41.6} & \textbf{67.8} & \textbf{25.7} & \underline{31.8} & \underline{18.2} & \textbf{26.9} & \textbf{32.3} & \textbf{50.5} & \underline{17.2} \\
\midrule
\multicolumn{14}{@{}l}{\tableheader{\textit{Student: Qwen3-4B-Base \enspace$\cdot$\enspace Teacher: Qwen3-32B}}} \\
\quad SFT ($\pref$) & 29.6 & 53.4 & \underline{39.8} & 62.0 & 35.6 & 61.4 & \underline{21.1} & 28.2 & 13.5 & 21.2 & 27.9 & 45.2 & -- \\
\quad Online OPD & \underline{32.6} & \underline{56.7} & 36.2 & \underline{63.6} & \textbf{42.0} & \underline{65.8} & 18.3 & \underline{31.2} & \underline{16.8} & \underline{24.5} & \underline{29.2} & \underline{48.4} & 146.0 \\
\quad \lopd{} & 28.1 & 50.2 & 26.8 & 57.5 & 32.6 & 60.1 & 9.2 & 22.9 & 11.2 & 19.8 & 21.6 & 42.1 & \textbf{13.6} \\
\rowcolor{gray!20} \quad \shortours{} & \textbf{33.4} & \textbf{57.2} & \textbf{41.6} & \textbf{66.3} & \underline{39.6} & \textbf{67.1} & \textbf{25.2} & \textbf{31.5} & \textbf{17.3} & \textbf{25.3} & \textbf{31.4} & \textbf{49.5} & \underline{13.8} \\
\bottomrule
\end{tabularx}
\caption{ Agentic search performance and training cost of a Qwen3-4B-Base student with two teachers. We report exact match (EM) and token-level F1 scores. GPU-h denotes training GPU-hours. Bold and underlined values indicate the best and second-best results among student methods in each column within each teacher block. Teacher results are provided for reference. }
\label{tab:main}
\end{table*}

\section{DistOPD: Distributional On-Policy Distillation}
\label{sec:method}

Building on this diagnosis, we introduce \shortours{} and evaluate whether compact distributional supervision improves agentic search while preserving offline training efficiency.

\subsection{Method}
\label{sec:method-loss}
To address the limitations of sampled-token supervision, we propose \ours{} (\shortours{}), which learns from compact distributional targets on fixed reference rollouts.
By retaining probability information over candidate tokens and aggregating the remaining mass into a residual token, \shortours{} provides supervision beyond the recorded token without caching full-vocabulary distributions.
Figure~\ref{fig:method} summarizes the framework.

\paragraph{Caching procedure.}
Whereas \lopd{} caches the teacher score of a single recorded token at each position, \shortours{} retains scores for a set of candidate tokens.
Let $\Sset_t^{ref}$ and $\Sset_t^T$ denote high-probability candidates from the frozen reference and teacher, respectively.
Together with the residual token $a_t^{res}$, they form

\begin{equation}
\Sset_t
=
\Sset_t^{ref}
\cup \Sset_t^T
\cup \{a_t^{res}\}.
\label{eq:candidate-set}
\end{equation}

Both reference and teacher candidates are collected in descending probability order until their cumulative mass reaches $\rho$, subject to a candidate-count cap.
The union retains reference-preferred tokens, adds teacher-supported alternatives, and ensures that the recorded token is included.

We aggregate all tokens outside $\Sset_t$ into a residual token and denote it as $a_t^{\mathrm{res}}$.
The residual token probability is

\begin{equation}
\pi(a_t^{\mathrm{res}}\mid s_t)
=
1-\sum_{j\in\Sset_t\setminus\{a_t^{\mathrm{res}}\}}
\pi(j\mid s_t).
\label{eq:coarsened}
\end{equation}

We cache reference and teacher log-probabilities for the candidate tokens, together with the their residual token masses as $\mathcal C_t=\Sset_t$.

This multi-candidate representation is used at positions selected according to the frozen reference distribution.
At the remaining positions, we set $\Sset_t=\{a_t, a_t^{res}\}$, yielding a two-cell partition containing the recorded token and its complementary residual token.
The selection rule, candidate-count limits, and scoring details are provided in Appendix~\ref{app:impl-cache}.

\paragraph{Target construction.}
At selected positions, we construct the target by geometrically interpolating the reference and teacher probabilities:

\begin{equation}
\hq_t(c)
=
\frac{\pref(c)^{1-\beta}\teacher(c)^\beta}{Z_t},
\qquad c\in\mathcal{C}_t.
\label{eq:dist-target}
\end{equation}

Here, $Z_t=\sum_{c'\in\mathcal{C}_t}\pref(c')^{1-\beta}\teacher(c')^\beta$ is the normalizing constant, and $\beta$ controls the teacher's contribution.

At unselected positions, we adjust the recorded-token weight using the teacher--reference gap
$\Delta_t=\log\teacher(a_t\mid s_t)-\log\pref(a_t\mid s_t)$:

\begin{equation}
\begin{aligned}
\hat{q}_{t}(c)
&=\frac{1}{Z_t}\\
&\quad{}\times
\begin{cases}
\pref(a_t\mid s_t)\,
e^{\beta\,\mathrm{clip}(\Delta_t,-10,10)},
& c=a_t,\\
1-\pref(a_t\mid s_t),
& c=a_t^{res}.
\end{cases}
\end{aligned}
\label{eq:two-cell-target}
\end{equation}

Here, $Z_t=\pref(a_t)e^{\beta\,\mathrm{clip}(\Delta_t,-10,10)}+\pref(a_t^{\mathrm{res}})$. The residual token weight retains its reference value before normalization.

\paragraph{Training objective.}
At every model-generated position, the student minimizes forward KL on the corresponding partition:

\begin{equation}
\ell_t(\theta)
=
\sum_{c\in\mathcal C_t}
\hq_{t}(c)
\log\frac{\hq_{t}(c)}{\ptheta(c\mid s_t)}.
\label{eq:dist-loss}
\end{equation}

For each candidate $j\in\Sset_t$, the gradient
$\partial\ell_t/\partial z_{t,j}
=\ptheta(j\mid s_t)-\hq_{t}(j)$
directly corrects deviations from its target probability.
Losses are averaged over model-generated positions within each trajectory and then over trajectories in the batch, without separate weighting of selected and unselected positions.
Training uses only cached supervision and requires no further teacher inference.

Under the state-wise assumptions of Theorem~\ref{thm:offline} and positive target cell probabilities, the loss is minimized when the student matches the target on every cell.
At selected positions, this constrains individual candidate-token probabilities and the aggregate residual token mass.
At unselected positions, it constrains only the recorded-token probability and its complementary mass.
Probabilities within the residual token remain unspecified.
Proofs and implementation details appear in Appendices~\ref{app:gradient} and~\ref{app:impl}, respectively.

\subsection{Experimental Results}
\label{sec:results-main}
\label{sec:results-cost}

We train \shortours{} models under the setup of Section~\ref{sec:exp-setup}, from the same cached rollouts as \lopd{}, and report the results in Table~\ref{tab:main}.

\paragraph{\shortours{} achieves performance comparable to online OPD.}
Table~\ref{tab:main} shows that \shortours{} improves upon the SFT reference policy $\pref$ on all five benchmarks with both teachers, whereas \lopd{} consistently falls below the same initialization.
This difference is particularly pronounced on 2WikiMultiHop and MuSiQue for Qwen3-32B.
On these benchmarks, \lopd{} reduces F1 relative to $\pref$ by $4.5$ and $5.3$ points, respectively, while \shortours{} improves it by $4.3$ and $3.3$ points.
The contrasting results show that offline distillation can improve upon the reference policy even in settings where scalar advantage regression leads to performance degradation.
Overall, \shortours{} achieves average F1 scores of $50.5$ for MiroThinker-1.7-mini and $49.5$ for Qwen3-32B, comparable to online OPD without requiring teacher evaluation during training.

\paragraph{\shortours{} reduces teacher--student mismatch.}
Following the diagnostics in Section~\ref{sec:analysis-regime}, we measure the student's agreement with the teacher and the concentration of its output distributions.
Relative to $\pref$, \shortours{} reduces the KL divergence to the teacher by $0.17$ and $0.18$ for MiroThinker-1.7-mini and Qwen3-32B, respectively.
Online OPD yields reductions of $0.06$ and $0.11$, whereas \lopd{} increases the divergence by $0.18$ and $0.46$.
The closer agreement achieved by \shortours{} is accompanied by only modest increases in next-token entropy of $0.11$ and $0.15$, compared with $1.2$ and $1.7$ for \lopd{}.
These measurements indicate that \shortours{} reduces teacher--student mismatch while producing substantially less flattening of the output distributions than scalar advantage regression.
Appendix~\ref{app:diagnosis} provides the metric definitions and detailed results.

\paragraph{\shortours{} preserves the training efficiency of \lopd{}.}
Although caching candidate distributions requires additional storage and distributional matching introduces extra computation, \shortours{} maintains a training cost comparable to \lopd{}.
With the Qwen3-32B teacher, \shortours{} requires $13.8$ GPU-hours, compared with $13.6$ for \lopd{} and $146.0$ for online OPD, reducing training GPU-hours by a factor of $10.6$ relative to online OPD.
The additional supervision increases the cache size to at most $4.2\times$ that of \lopd{}, while keeping it below $8$\,GiB.
As with \lopd{}, the cache is constructed once and can be reused across training runs using the same cached data.
Appendix~\ref{app:cost} provides the detailed cost breakdown.

\begin{figure}[t]
\centering
\includegraphics[width=0.9\columnwidth]{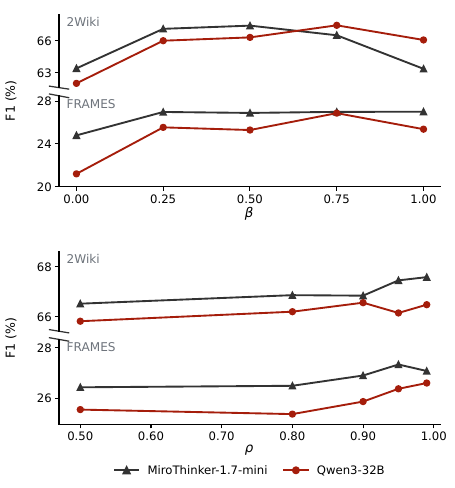}
\caption{\textbf{\shortours{} is relatively stable across intermediate $\beta$ values and the tested $\rho$ range.} F1 (\%) is shown for both teachers as $\beta$ (top) and $\rho$ (bottom) vary; $\beta{=}0$ denotes the SFT baseline. At $\beta{=}1$, MiroThinker-1.7-mini shows a marked drop on 2WikiMultiHop. Broken axes separate the two benchmarks.}
\label{fig:ablation}
\end{figure}

\begin{table}[t]
\centering\small
\setlength{\tabcolsep}{4pt}
\begin{tabular}{@{}lcccc@{}}
\toprule
\tableheader{Method} & \tableheader{AIME~24} & \tableheader{AIME~25} & \tableheader{Avg.} & \tableheader{GPU-h}\\
\midrule
SFT ($\pref$) & 55.4 & 48.6 & 52.0 & --\\
Online OPD & 62.7 & 54.1 & 58.4 & 94.0\\
\lopd{} & 63.1 & 55.3 & 59.2 & 8.0\\
\rowcolor{gray!20} \shortours{} & \textbf{64.1} & \textbf{56.2} & \textbf{60.2} & 8.2\\
\bottomrule
\end{tabular}
\caption{\textbf{\shortours{} generalizes to mathematical reasoning with competitive performance and low training cost.} Results use Qwen3-4B-Base and Qwen3-32B as student and teacher, following \citet{wu2026lightning}. Scores are pass@1 (\%) estimated from $32$ samples per problem; bold marks the best scores. GPU-h excludes one-time cache construction for offline methods.}
\label{tab:math}
\end{table}

\section{Analysis}
\label{sec:further}
Beyond the main results on agentic search, we investigate the design choices and broader applicability of \shortours{}.
Our analysis examines its sensitivity to hyperparameter choices, its generalization to mathematical reasoning, and the role of KL direction in distributional supervision.
These experiments assess whether its effectiveness extends across configurations and task domains, and clarify how the training objective influences performance.

\paragraph{Q1: How do the DistOPD hyperparameters $\beta$ and $\rho$ affect performance?}
\label{sec:ablation}
\label{sec:results-ablation}

We vary $\beta$ and $\rho$ independently while keeping all other settings fixed, and evaluate on 2WikiMultiHop and FRAMES with both teachers.
As shown in Figure~\ref{fig:ablation}, performance remains relatively stable for $\beta\in[0.25,0.75]$, suggesting that effective interpolation does not require a precisely tuned coefficient.
This stability does not extend to every endpoint: at $\beta=1$, performance drops on 2WikiMultiHop with MiroThinker-1.7-mini.

For $\rho$, performance varies modestly across the tested range from $0.5$ to $0.99$, and all settings outperform \lopd{}.
Increasing the threshold does not consistently improve F1, with $\rho=0.5$ already retaining gains on both benchmarks with both teachers.
This indicates that a high teacher candidate-selection threshold is not necessary to obtain these improvements when the reference candidates and residual token are retained.

\paragraph{Q2: Can DistOPD generalize to other domains?}
\label{sec:analysis-math}
To assess whether \shortours{} generalizes beyond agentic search, we evaluate it on mathematical reasoning following the same setup with \citet{wu2026lightning}.
Following the same training data, budget, and evaluation protocol of \citet{wu2026lightning}, we report average pass@1 on AIME~24 and AIME~25 (Table~\ref{tab:math}).
\lopd{} achieves an average score of $59.2$, compared with $58.4$ for online OPD.
As shown in Table~\ref{tab:math}, \shortours{} achieves the highest pass@1 on both AIME~24 and AIME~25, averaging $60.2$, compared with $59.2$ for \lopd{} and $58.4$ for online OPD.
With only $30$ problems per benchmark, we interpret these modest gains conservatively as evidence of comparable performance.
\shortours{} requires $8.2$ training GPU-hours, comparable to \lopd{} ($8.0$) and approximately one-eleventh of online OPD ($94.0$).
These results support the generalization of \shortours{} to mathematical reasoning, preserving both distillation performance and offline training efficiency.

\paragraph{Q3: How does the KL direction affect DistOPD performance?}
\label{sec:analysis-kl}

We compare forward and reverse KL on 2WikiMultiHop and FRAMES with the Qwen3-32B teacher.
The two objectives achieve similar performance: forward KL obtains F1 scores of $66.3$ and $25.3$, respectively, while reverse KL obtains $65.8$ and $24.7$.
Both outperform \lopd{} on the two benchmarks, indicating that the improvements are retained under either KL direction.
These results suggest that the improvements over \lopd{} are not specific to the KL direction in the evaluated settings.
We adopt forward KL because its gradient for each candidate logit is the student--target probability difference, providing a direct corrective signal for alternatives underweighted by the student.
This choice aligns with our aim of explicitly supervising probability allocation among candidate tokens.

\section{Related Work}
\label{sec:related}

\paragraph{On-Policy Distillation.}
OPD trains a student on its own generated sequences using teacher feedback \citep{agarwal2024gkd,gu2024minillm}.
A common approach updates the student using the difference between teacher and student log-probabilities for each generated token~\citep{lu2025opd}.
To avoid querying the teacher repeatedly during training, \lopd{}~\citep{wu2026lightning,wu2026lightning2} scores rollouts from a frozen reference policy once and trains on cached teacher log-probabilities.
This offline approach achieves performance comparable to online OPD on mathematical reasoning and code generation at substantially lower training cost.
Our work extends this investigation to agentic search, identifying limitations of scalar advantage regression on fixed reference rollouts.
Building on the same offline setting, \shortours{} augments sampled-token supervision with compact distributional targets at selected positions.

\paragraph{Agentic Search.}
Agentic search integrates reasoning with iterative retrieval, allowing language models to acquire external information as a task unfolds \citep{react,searcho1}.
Search-R1 and R1-Searcher learn such behavior through reinforcement learning with outcome-based rewards \citep{searchr1,r1searcher}.
A complementary line of work transfers agent capabilities from teachers to smaller students through correction of student-generated trajectories \citep{lyu2026score} or reuse of teacher-generated tool modules \citep{kang2025agentdistill}.
More closely related to our work, recent approaches adapt OPD to multi-turn interactions through temporal curricula \citep{wang2026tcod,guidedopd2026,liu2026asp} and step-wise reweighting of teacher supervision \citep{zhong2026sod}.
These multi-turn OPD methods obtain teacher feedback on trajectories collected during training.
Our work studies a complementary offline setting in which reference rollouts and teacher supervision are fixed before optimization.

\paragraph{Distillation Objectives and Supervision.}
Language-model distillation uses divergence objectives to align teacher and student output distributions~\citep{wen2023fdistill,gu2024minillm,agarwal2024gkd}.
TAID further constructs intermediate targets by interpolating student and teacher logits~\citep{shing2025taid}.
Our work studies compact distributional supervision for offline OPD, using cached probabilities over candidate tokens and a residual bucket.
Unlike TAID, \shortours{} interpolates between frozen reference and teacher distributions, keeping the targets fixed throughout optimization.

\section{Conclusion}
\label{sec:conclusion}

We study offline on-policy distillation in multi-turn agentic search, where \lopd{} falls below its SFT initialization although the same recipe is effective on mathematical reasoning.
Our analysis links this failure to the larger teacher--reference gap of agentic search: since a cached scalar constrains only the probability of the recorded token, the student can satisfy it by flattening its distribution rather than moving toward the teacher.
\shortours{} keeps the frozen rollouts and training cost of \lopd{} but caches a few candidate probabilities per position, which is enough to improve upon SFT, match online OPD, and bring the student closer to the teacher.
Offline distillation can therefore remain effective in agentic settings, as long as the cache retains part of the teacher distribution rather than a single scalar per token.

\section*{Limitations}
\label{sec:limitations}

Our experiments use a single 4B student, which limits the scope of our empirical findings.
Further evaluation is needed to establish whether the observed limitations of \lopd{} and the benefits of \shortours{} extend to other student models and scales.
Our theoretical analysis focuses on logit contraction as one possible mechanism consistent with the observed distributional flattening.
Whether scalar supervision on fixed rollouts admits other forms of degeneration remains an open question.

\bibliography{custom}

@inproceedings{agarwal2024gkd,
  author    = {Agarwal, Rishabh and Vieillard, Nino and Zhou, Yongchao and Stanczyk, Piotr and Ramos, Sabela and Geist, Matthieu and Bachem, Olivier},
  title     = {On-Policy Distillation of Language Models: Learning from Self-Generated Mistakes},
  booktitle = {The Twelfth International Conference on Learning Representations ({ICLR})},
  year      = {2024},
  url       = {https://arxiv.org/abs/2306.13649}
}

@inproceedings{gu2024minillm,
  author    = {Gu, Yuxian and Dong, Li and Wei, Furu and Huang, Minlie},
  title     = {{M}ini{LLM}: Knowledge Distillation of Large Language Models},
  booktitle = {The Twelfth International Conference on Learning Representations ({ICLR})},
  year      = {2024},
  url       = {https://arxiv.org/abs/2306.08543}
}

@article{hinton2015distilling,
  author  = {Hinton, Geoffrey and Vinyals, Oriol and Dean, Jeff},
  title   = {Distilling the Knowledge in a Neural Network},
  journal = {arXiv preprint},
  volume  = {arXiv:1503.02531},
  year    = {2015},
  url     = {https://arxiv.org/abs/1503.02531},
  note    = {NIPS 2014 Deep Learning Workshop}
}

@article{lu2025opd,
  author  = {Lu, Kevin and {Thinking Machines Lab}},
  title   = {On-Policy Distillation},
  journal = {Thinking Machines Lab: Connectionism},
  year    = {2025},
  doi     = {10.64434/tml.20251026},
  url     = {https://thinkingmachines.ai/blog/on-policy-distillation}
}

@article{wu2026lightning,
  author  = {Wu, Yecheng and Han, Song and Cai, Han},
  title   = {Lightning {OPD}: Efficient Post-Training for Large Reasoning Models with Offline On-Policy Distillation},
  journal = {arXiv preprint},
  volume  = {arXiv:2604.13010},
  year    = {2026},
  url     = {https://arxiv.org/abs/2604.13010}
}

@article{wu2026lightning2,
  author  = {Wu, Yecheng and Han, Song and Cai, Han},
  title   = {Lightning {OPD} 2.0: Mitigating Style Bias in Cross-Teacher On-Policy Distillation for Large Reasoning Models},
  journal = {arXiv preprint},
  volume  = {arXiv:2607.28449},
  year    = {2026},
  url     = {https://arxiv.org/abs/2607.28449}
}

@article{asyncopd2026,
  author  = {Kang, Wonjun and Galim, Kevin and Oh, Seunghyuk and Kang, Minjun and Park, Sanghyun and Kim, Donghoon and Lee, Minjae and Kim, Minseo and Tiwari, Rishabh and Zeng, Yuchen and Koo, Hyung Il and Lee, Kangwook},
  title   = {{A}sync{OPD}: How Stale Can On-Policy Distillation Be?},
  journal = {arXiv preprint},
  volume  = {arXiv:2606.24143},
  year    = {2026},
  url     = {https://arxiv.org/abs/2606.24143}
}

@article{fu2026revisiting,
  author  = {Fu, Yuqian and Huang, Haohuan and Jiang, Kaiwen and Liu, Jiacai and Jiang, Zhuo and Zhu, Yuanheng and Zhao, Dongbin},
  title   = {Revisiting On-Policy Distillation: Empirical Failure Modes and Simple Fixes},
  journal = {arXiv preprint},
  volume  = {arXiv:2603.25562},
  year    = {2026},
  url     = {https://arxiv.org/abs/2603.25562}
}

@inproceedings{jang2026veto,
  author    = {Jang, Ijun and Yeom, Jewon and Yeo, Juan and Lim, Hyunggyu and Kim, Taesup},
  title     = {Stable On-Policy Distillation through Adaptive Target Reformulation},
  booktitle = {Findings of the {A}ssociation for {C}omputational {L}inguistics: {ACL} 2026},
  pages     = {42217--42227},
  publisher = {Association for Computational Linguistics},
  address   = {San Diego, California, United States},
  month     = jul,
  year      = {2026},
  doi       = {10.18653/v1/2026.findings-acl.2094},
  url       = {https://aclanthology.org/2026.findings-acl.2094/}
}

@article{jin2026eopd,
  author  = {Jin, Woogyeol and Min, Taywon and Yang, Yongjin and Wei, Dennis and Zhou, Yi and Kadhe, Swanand Ravindra and Baracaldo, Nathalie and Lee, Kimin},
  title   = {Entropy-Aware On-Policy Distillation of Language Models},
  journal = {arXiv preprint},
  volume  = {arXiv:2603.07079},
  year    = {2026},
  url     = {https://arxiv.org/abs/2603.07079}
}

@article{ko2026relaxed,
  author  = {Ko, Jongwoo and Abdali, Sara and Kim, Young Jin and Chen, Tianyi and Cameron, Pashmina},
  title   = {Scaling Reasoning Efficiently via Relaxed On-Policy Distillation},
  journal = {arXiv preprint},
  volume  = {arXiv:2603.11137},
  year    = {2026},
  url     = {https://arxiv.org/abs/2603.11137}
}

@inproceedings{wen2023fdistill,
  author    = {Wen, Yuqiao and Li, Zichao and Du, Wenyu and Mou, Lili},
  title     = {f-Divergence Minimization for Sequence-Level Knowledge Distillation},
  booktitle = {Proceedings of the 61st Annual Meeting of the Association for Computational Linguistics (Volume 1: Long Papers)},
  pages     = {10817--10834},
  publisher = {Association for Computational Linguistics},
  address   = {Toronto, Canada},
  month     = jul,
  year      = {2023},
  doi       = {10.18653/v1/2023.acl-long.605},
  url       = {https://aclanthology.org/2023.acl-long.605/}
}

@article{searchr1,
  author  = {Jin, Bowen and Zeng, Hansi and Yue, Zhenrui and Yoon, Jinsung and Arik, Sercan and Wang, Dong and Zamani, Hamed and Han, Jiawei},
  title   = {{S}earch-{R}1: Training {LLM}s to Reason and Leverage Search Engines with Reinforcement Learning},
  journal = {arXiv preprint},
  volume  = {arXiv:2503.09516},
  year    = {2025},
  url     = {https://arxiv.org/abs/2503.09516}
}

@article{r1searcher,
  author  = {Song, Huatong and Jiang, Jinhao and Min, Yingqian and Chen, Jie and Chen, Zhipeng and Zhao, Wayne Xin and Fang, Lei and Wen, Ji-Rong},
  title   = {{R}1-{S}earcher: Incentivizing the Search Capability in {LLM}s via Reinforcement Learning},
  journal = {arXiv preprint},
  volume  = {arXiv:2503.05592},
  year    = {2025},
  url     = {https://arxiv.org/abs/2503.05592}
}

@article{wang2026tcod,
  author  = {Wang, Jiaqi and Zhang, Wenhao and Shi, Weijie and Li, Yaliang and Cheng, James},
  title   = {{TCOD}: Exploring Temporal Curriculum in On-Policy Distillation for Multi-turn Autonomous Agents},
  journal = {arXiv preprint},
  volume  = {arXiv:2604.24005},
  year    = {2026},
  url     = {https://arxiv.org/abs/2604.24005}
}

@article{zhong2026sod,
  author  = {Zhong, Qiyong and Zheng, Mao and Song, Mingyang and Lin, Xin and Sun, Jie and Jiang, Houcheng and Wang, Xiang and Fang, Junfeng},
  title   = {{SOD}: Step-wise On-policy Distillation for Small Language Model Agents},
  journal = {arXiv preprint},
  volume  = {arXiv:2605.07725},
  year    = {2026},
  url     = {https://arxiv.org/abs/2605.07725}
}

@article{guidedopd2026,
  author  = {Li, Gengsheng and Zheng, Mao and Song, Mingyang and Liu, Ruiqi and Yang, Tianyu and Sun, Jie and Zhong, Qiyong and Guo, Haiyun and Fang, Junfeng and Zhang, Dan and Wang, Jinqiao},
  title   = {On-Policy Distillation with Curriculum Turn-level Guidance for Multi-turn Agents},
  journal = {arXiv preprint},
  volume  = {arXiv:2606.15912},
  year    = {2026},
  url     = {https://arxiv.org/abs/2606.15912}
}

@inproceedings{hotpotqa,
  author    = {Yang, Zhilin and Qi, Peng and Zhang, Saizheng and Bengio, Yoshua and Cohen, William and Salakhutdinov, Ruslan and Manning, Christopher D.},
  title     = {{H}otpot{QA}: A Dataset for Diverse, Explainable Multi-hop Question Answering},
  booktitle = {Proceedings of the 2018 Conference on Empirical Methods in Natural Language Processing},
  pages     = {2369--2380},
  publisher = {Association for Computational Linguistics},
  address   = {Brussels, Belgium},
  month     = oct # "-" # nov,
  year      = {2018},
  doi       = {10.18653/v1/D18-1259},
  url       = {https://aclanthology.org/D18-1259/}
}

@inproceedings{twowikimultihopqa,
  author    = {Ho, Xanh and Duong Nguyen, Anh-Khoa and Sugawara, Saku and Aizawa, Akiko},
  title     = {Constructing A Multi-hop {QA} Dataset for Comprehensive Evaluation of Reasoning Steps},
  booktitle = {Proceedings of the 28th International Conference on Computational Linguistics},
  pages     = {6609--6625},
  publisher = {International Committee on Computational Linguistics},
  address   = {Barcelona, Spain (Online)},
  month     = dec,
  year      = {2020},
  doi       = {10.18653/v1/2020.coling-main.580},
  url       = {https://aclanthology.org/2020.coling-main.580/}
}

@article{musique,
  author    = {Trivedi, Harsh and Balasubramanian, Niranjan and Khot, Tushar and Sabharwal, Ashish},
  title     = {{M}u{S}i{Q}ue: Multihop Questions via Single-hop Question Composition},
  journal   = {Transactions of the Association for Computational Linguistics},
  volume    = {10},
  pages     = {539--554},
  publisher = {MIT Press},
  address   = {Cambridge, MA},
  year      = {2022},
  doi       = {10.1162/tacl_a_00475},
  url       = {https://aclanthology.org/2022.tacl-1.31/}
}

@inproceedings{bamboogle,
  author    = {Press, Ofir and Zhang, Muru and Min, Sewon and Schmidt, Ludwig and Smith, Noah A. and Lewis, Mike},
  title     = {Measuring and Narrowing the Compositionality Gap in Language Models},
  booktitle = {Findings of the Association for Computational Linguistics: {EMNLP} 2023},
  pages     = {5687--5711},
  publisher = {Association for Computational Linguistics},
  address   = {Singapore},
  month     = dec,
  year      = {2023},
  doi       = {10.18653/v1/2023.findings-emnlp.378},
  url       = {https://aclanthology.org/2023.findings-emnlp.378/}
}

@inproceedings{frames,
  author    = {Krishna, Satyapriya and Krishna, Kalpesh and Mohananey, Anhad and Schwarcz, Steven and Stambler, Adam and Upadhyay, Shyam and Faruqui, Manaal},
  title     = {Fact, Fetch, and Reason: A Unified Evaluation of Retrieval-Augmented Generation},
  booktitle = {Proceedings of the 2025 Conference of the Nations of the Americas Chapter of the Association for Computational Linguistics: Human Language Technologies (Volume 1: Long Papers)},
  pages     = {4745--4759},
  publisher = {Association for Computational Linguistics},
  address   = {Albuquerque, New Mexico},
  month     = apr,
  year      = {2025},
  doi       = {10.18653/v1/2025.naacl-long.243},
  url       = {https://aclanthology.org/2025.naacl-long.243/}
}

@article{qwen3,
  author  = {Yang, An and Li, Anfeng and Yang, Baosong and others},
  title   = {{Q}wen3 Technical Report},
  journal = {arXiv preprint},
  volume  = {arXiv:2505.09388},
  year    = {2025},
  url     = {https://arxiv.org/abs/2505.09388}
}

@article{mirothinker,
  author  = {{MiroMind Team} and Bai, Song and Bing, Lidong and Chen, Carson and Chen, Guanzheng and Chen, Yuntao and others},
  title   = {{M}iro{T}hinker: Pushing the Performance Boundaries of Open-Source Research Agents via Model, Context, and Interactive Scaling},
  journal = {arXiv preprint},
  volume  = {arXiv:2511.11793},
  year    = {2025},
  url     = {https://arxiv.org/abs/2511.11793}
}

@inproceedings{wiki18dpr,
  author    = {Karpukhin, Vladimir and Oguz, Barlas and Min, Sewon and Lewis, Patrick and Wu, Ledell and Edunov, Sergey and Chen, Danqi and Yih, Wen-tau},
  title     = {Dense Passage Retrieval for Open-Domain Question Answering},
  booktitle = {Proceedings of the 2020 Conference on Empirical Methods in Natural Language Processing ({EMNLP})},
  pages     = {6769--6781},
  publisher = {Association for Computational Linguistics},
  address   = {Online},
  month     = nov,
  year      = {2020},
  doi       = {10.18653/v1/2020.emnlp-main.550},
  url       = {https://aclanthology.org/2020.emnlp-main.550/}
}

@article{e5,
  author  = {Wang, Liang and Yang, Nan and Huang, Xiaolong and Jiao, Binxing and Yang, Linjun and Jiang, Daxin and Majumder, Rangan and Wei, Furu},
  title   = {Text Embeddings by Weakly-Supervised Contrastive Pre-training},
  journal = {arXiv preprint},
  volume  = {arXiv:2212.03533},
  year    = {2022},
  url     = {https://arxiv.org/abs/2212.03533}
}

@misc{liu2026asp,
      title = {Search, Do not Guess: Teaching Small Language Models to Be Effective Search Agents},
      author = {Yizhou Liu and Qi Sun and Yulin Chen and Siyue Zhang and Chen Zhao},
      year = {2026},
      eprint = {2604.04651},
      archivePrefix = {arXiv},
      primaryClass = {cs.AI},
      url = {https://arxiv.org/abs/2604.04651}
}

@article{searcho1,
  title        = {Search-o1: Agentic Search-Enhanced Large Reasoning Models},
  author       = {Li, Xiaoxi and Dong, Guanting and Jin, Jiajie and Zhang, Yuyao and
                  Zhou, Yujia and Zhu, Yutao and Zhang, Peitian and Dou, Zhicheng},
  journal      = {arXiv preprint arXiv:2501.05366},
  year         = {2025},
  eprint       = {2501.05366},
  archivePrefix= {arXiv},
  primaryClass = {cs.CL}
}

@inproceedings{react,
  title        = {{ReAct}: Synergizing Reasoning and Acting in Language Models},
  author       = {Yao, Shunyu and Zhao, Jeffrey and Yu, Dian and Du, Nan and
                  Shafran, Izhak and Narasimhan, Karthik and Cao, Yuan},
  booktitle    = {International Conference on Learning Representations (ICLR)},
  year         = {2023},
  eprint       = {2210.03629},
  archivePrefix= {arXiv},
  primaryClass = {cs.CL}
}

@inproceedings{kang2025agentdistill,
  title        = {Distilling {LLM} Agent into Small Models with Retrieval and Code Tools},
  author       = {Kang, Minki and Jeong, Jongwon and Lee, Seanie and Cho, Jaewoong and
                  Hwang, Sung Ju},
  booktitle    = {Advances in Neural Information Processing Systems (NeurIPS)},
  year         = {2025},
  note         = {Spotlight},
  eprint       = {2505.17612},
  archivePrefix= {arXiv},
  primaryClass = {cs.CL}
}

@inproceedings{lyu2026score,
  title        = {Student-Centered Distillation Narrows the Agentic Gap Between Small
                  and Large {LLM}s},
  author       = {Lyu, Yuanjie and Wang, Chengyu and Huang, Jun and Xu, Tong},
  booktitle    = {Proceedings of the 43rd International Conference on Machine Learning
                  (ICML)},
  year         = {2026},
  eprint       = {2509.14257},
  archivePrefix= {arXiv},
  primaryClass = {cs.CL}
}

@misc{shing2025taid,
      title={TAID: Temporally Adaptive Interpolated Distillation for Efficient Knowledge Transfer in Language Models}, 
      author={Makoto Shing and Kou Misaki and Han Bao and Sho Yokoi and Takuya Akiba},
      year={2025},
      eprint={2501.16937},
      archivePrefix={arXiv},
      primaryClass={cs.LG},
      url={https://arxiv.org/abs/2501.16937}, 
}

\newpage

\appendix

\section{Theoretical Analysis of Scalar and Distributional Supervision}
\label{app:proofs}

This appendix provides the derivations supporting the theoretical analysis and the training objective of \shortours{}.
We first characterize the constraints imposed by scalar advantage regression on fixed reference rollouts and establish conditions under which logit contraction satisfies these constraints.
We then derive the cached distributional target, its gradients, and the constraints imposed by candidate-token and bucket supervision.
These results characterize state-wise objectives and their minimizers; they do not establish the optimization trajectories or downstream performance of models with shared parameters.

\subsection{Notation and Preliminaries}
\label{app:softmax}

\providecommand{\ind}{\mathbf 1}

We fix one model-generated token position and omit the index $t$.
Let $\mathcal V$ be a vocabulary of size $V\ge2$, $y\in\mathcal V$ the recorded token, and $z\in\R^V$ the student's finite logits.
The student distribution is $p=\mathrm{softmax}(z)$:

\begin{equation}
p_j=\frac{e^{z_j}}{\sum_k e^{z_k}}>0.
\label{eq:app-softmax}
\end{equation}

The teacher and reference distributions are denoted by $q$ and $\pi$, respectively, and both have full support.
They are fixed during optimization; only the probabilities required by the training objective are cached.
We write $c=\log q_y<0$ and define

\begin{equation}
A=\log q_y-\log p_y=c-\log p_y.
\label{eq:app-adv}
\end{equation}

The teacher term is fixed, while the student term and hence $A$ are recomputed during optimization.
We use $e_j$ for the $j$-th basis vector, $\mathbf 1$ for the all-ones vector, and $\Delta$ for the probability simplex.
Since $\mathrm{softmax}(z+s\mathbf 1)=\mathrm{softmax}(z)$, a probability distribution determines its logits only up to a common additive shift.

Results concerning state-wise minimizers assume unrestricted logits parameterized independently at each analyzed context, with one recorded continuation per context.
The gradient identities themselves do not require independent parameterization.
All sums over trajectory positions include only model-generated tokens; retrieved content remains part of the conditioning contexts.

\begin{lemma}[Softmax log-gradient]
\label{lem:softmax}
For every $i,j$,

\begin{equation}
\frac{\partial\log p_j}{\partial z_i}
=\ind[i=j]-p_i,
\qquad
\frac{\partial p_j}{\partial z_i}
=p_j\big(\ind[i=j]-p_i\big).
\end{equation}

Equivalently, $\nabla_z\log p_j=e_j-p$.
\end{lemma}

\begin{proof}
Write $\log p_j=z_j-\log\sum_k e^{z_k}$ and differentiate.
The second identity follows from
$\nabla_zp_j=p_j\nabla_z\log p_j$.
\end{proof}

\begin{corollary}[Group probabilities]
\label{cor:group}
For a nonempty set $G\subseteq\mathcal V$, let $p_G=\sum_{k\in G}p_k$.
Then

\begin{equation}
\frac{\partial\log p_G}{\partial z_i}
=\ind[i\in G]\frac{p_i}{p_G}-p_i.
\end{equation}

\end{corollary}

\begin{proof}
By Lemma~\ref{lem:softmax},

\[
\frac{\partial p_G}{\partial z_i}
=\ind[i\in G]p_i-p_ip_G.
\]

Dividing by $p_G>0$ gives the result.
\end{proof}

\subsection{Scalar Advantage Regression and Its Minimizers}
\label{app:regression}

\begin{proposition}[Offline advantage as regression]
\label{prop:regression}
Let $\mathcal D$ be a finite dataset of trajectories sampled from $\pref$ and held fixed.
At each recorded position, define
$A_t(\theta)=\log\teacher(y_t\mid s_t)-\log\ptheta(y_t\mid s_t)$.
Let

\begin{align}
J_{\mathrm{off}}(\theta)
&=\E_{\tau\sim\mathcal D}\Big[
\textstyle\sum_t A_t(\theta)
\nabla_\theta\log\ptheta(y_t\mid s_t)
\Big],
\label{eq:app-goff}\\
L(\theta)
&=\E_{\tau\sim\mathcal D}\Big[
\textstyle\sum_t\tfrac12 A_t(\theta)^2
\Big],
\label{eq:app-L}
\end{align}

where the expectations denote empirical averages.
Then $J_{\mathrm{off}}(\theta)=-\nabla_\theta L(\theta)$.
In the update, $A_t(\theta)$ is used as a detached weight; in $L(\theta)$, its dependence on $\theta$ is differentiated.
\end{proposition}

\begin{proof}
The empirical average has $\theta$-independent weights, so it suffices to compare individual terms.
Since the teacher log-probability is fixed,

\[
\nabla_\theta\tfrac12 A_t(\theta)^2
=-A_t(\theta)\nabla_\theta
\log\ptheta(y_t\mid s_t).
\]

Summing over the fixed dataset establishes the identity.
\end{proof}

\begin{remark}[Fixed sampling]
\label{rem:freezing}
The regression equivalence relies on a sampling distribution that is independent of the current parameters.
With fresh samples from the student, the expected detached-advantage update instead equals the negative reverse-KL gradient at a fixed context, as shown in Proposition~\ref{prop:online}.
This update is not the negative gradient of the expected squared advantage under the current student distribution.
\end{remark}

\begin{remark}[Advantage clipping]
\label{rem:clipping}
Let $h_\tau$ be the Huber penalty with
$h_\tau'(r)=\mathrm{clip}(r,-\tau,\tau)$.
Then

\[
\nabla_\theta h_\tau(A_t)
=-\mathrm{clip}(A_t,-\tau,\tau)
\nabla_\theta\log\ptheta(y_t\mid s_t).
\]

Thus, clipping the advantage at $\tau=10$ replaces the squared penalty with a Huber penalty.
Both penalties are minimized precisely at zero residual, so they have the same state-wise minimizers.
For shared model parameters, their global minimizer sets need not coincide if all residuals cannot be made zero simultaneously.
\end{remark}

\label{app:manifold}
At a single position, the squared loss is

\begin{equation}
L_{\mathrm{pos}}(z)
=\tfrac12\big(c-\log p_y(z)\big)^2.
\label{eq:app-Lpos}
\end{equation}

\begin{proposition}[Minimizer manifold]
\label{prop:manifold}
The following statements hold.
\begin{enumerate}
\item[(i)]
$\nabla_zL_{\mathrm{pos}}(z)=-A(e_y-p)$.
Since $e_y-p\neq0$, a finite-logit stationary point satisfies $A=0$ and is a global minimizer.
\item[(ii)]
The minimizer set
$M_c=\{z:\log p_y(z)=c\}$
is a nonempty smooth submanifold of $\R^V$ of dimension $V-1$.
It can be written as

\begin{equation}
z_y
=c-\log(1-e^c)
+\log\textstyle\sum_{j\neq y}e^{z_j}.
\label{eq:app-graph}
\end{equation}

\item[(iii)]
In probability space, the minimizers satisfy $p_y=q_y$ and retain $V-2$ degrees of freedom.
\end{enumerate}
\end{proposition}

\begin{proof}
Part (i) follows from Lemma~\ref{lem:softmax} and the chain rule.
Because $p_y<1$, the vector $e_y-p$ is nonzero, so stationarity is equivalent to $A=0$.

For part (ii), exponentiating $\log p_y=c$ gives

\[
e^{z_y}(1-e^c)
=e^c\sum_{j\neq y}e^{z_j}.
\]

Since $0<e^c<1$, taking logarithms yields Equation~\eqref{eq:app-graph}.
Every choice of the $V-1$ other logits determines exactly one $z_y$, giving a smooth graph over $\R^{V-1}$.

For part (iii), the probability simplex has dimension $V-1$.
The constraint $p_y=q_y\in(0,1)$ fixes one coordinate, leaving the positive probabilities of the remaining tokens subject only to their sum being $1-q_y$.
Their feasible set therefore has dimension $V-2$.
\end{proof}

\begin{remark}[Supervision over alternatives]
\label{rem:manifold}
The scalar objective fixes the recorded-token probability but does not specify the relative probabilities of alternative tokens.
For $V\ge3$, different minimizers can rank alternatives differently while attaining the same zero loss.
This underdetermination concerns the available teacher constraints, rather than an absence of gradients on alternative logits.
\end{remark}

\subsection{Logit-Contracted Minimizers}
\label{app:scaling}

\begin{proposition}[The rescaling solution]
\label{prop:scaling}
Fix logits $z$ for which $y$ is the unique highest-logit token.
Let $m_j=z_y-z_j>0$ for $j\neq y$ and define
$f(\alpha)=\log p_y(\alpha z)$ for $\alpha\ge0$.
Then
\begin{enumerate}
\item[(i)]
$f(\alpha)=-\log\big(1+\sum_{j\neq y}e^{-\alpha m_j}\big)$;
\item[(ii)]
$f'(\alpha)=z_y-\E_{p_{\alpha z}}[z]>0$;
\item[(iii)]
$f(0)=-\log V$ and
$\lim_{\alpha\to\infty}f(\alpha)=0$.
\end{enumerate}
Consequently, for every $c\in[-\log V,0)$, there is a unique $\alpha^\star\ge0$ such that $\alpha^\star z\in M_c$.
In particular, if

\begin{equation}
\frac1V<q_y<p_y(z),
\label{eq:app-contraction-condition}
\end{equation}

then $\alpha^\star\in(0,1)$.
This contraction preserves token rankings and strictly increases the entropy of the student distribution.
\end{proposition}

\begin{proof}
Dividing the numerator and denominator of $p_y(\alpha z)$ by $e^{\alpha z_y}$ gives part (i).
Differentiating
$f(\alpha)=\alpha z_y-\log\sum_k e^{\alpha z_k}$
gives part (ii).
The expectation of $z$ under the full-support distribution $p_{\alpha z}$ is strictly smaller than its unique maximum $z_y$.
Part (iii) follows by evaluating at $\alpha=0$ and taking the limit.
Continuity and strict monotonicity establish existence and uniqueness.
Under Equation~\eqref{eq:app-contraction-condition},
$f(0)<\log q_y<f(1)$, so $\alpha^\star\in(0,1)$.

Positive rescaling preserves all logit orderings.
For entropy,

\[
H(p_{\alpha z})
=\log\sum_j e^{\alpha z_j}
-\alpha\,\E_{p_{\alpha z}}[z].
\]

Differentiation yields

\begin{equation}
\frac{d}{d\alpha}H(p_{\alpha z})
=-\alpha\,\operatorname{Var}_{p_{\alpha z}}(z)<0
\qquad(\alpha>0),
\label{eq:app-entropy-scaling}
\end{equation}

because $z$ is nonconstant.
Therefore, reducing $\alpha$ from $1$ to $\alpha^\star$ strictly increases entropy.
\end{proof}

\begin{remark}[Interpretation of contraction]
\label{rem:scaling}
Under the stated conditions, a temperature change alone can satisfy the recorded-token constraint without matching the teacher's relative preferences among alternatives.
The resulting distribution may or may not equal the teacher distribution.
The proposition establishes the existence of a logit-contracted minimizer; it does not imply that optimization follows this rescaling path or that the resulting policy has lower task performance.
\end{remark}

\label{app:sign}
\begin{proposition}[Mean advantage under reference sampling]
\label{prop:gibbs}
For full-support distributions $p,q$,

\[
\E_{y\sim p}[\log q_y-\log p_y]
=-\KL(p\|q)\le0,
\]

with equality if and only if $p=q$.
At initialization, where $p=\pi$, this gives
$\E_{y\sim\pi}[A]=-\KL(\pi\|q)$.
For a general current student $p$ under reference sampling,

\begin{equation}
\E_{y\sim\pi}[A]
=\KL(\pi\|p)-\KL(\pi\|q).
\label{eq:app-reference-mean}
\end{equation}

\end{proposition}

\begin{proof}
The first identity follows from the definition of KL divergence, and its sign follows from Gibbs' inequality.
For the final identity, write

\[
\log\frac{q_y}{p_y}
=\log\frac{\pi_y}{p_y}
-\log\frac{\pi_y}{q_y}
\]

and take the expectation under $\pi$.
\end{proof}

\begin{remark}[Relation to the initial gap]
At initialization, a negative recorded-token gap is equivalent to $q_y<\pi_y$.
It supplies one of the probability inequalities required by Proposition~\ref{prop:scaling}, but does not establish the argmax condition or the lower bound $q_y>1/V$.
A negative mean gap likewise does not establish these conditions at individual positions.
The expectations above are population expectations at a fixed context, not identities for every finite empirical sample.
\end{remark}

\subsection{The Online Update at a Fixed Context}
\label{app:online}

For the stationary-distribution analysis, we consider the unclipped online OPD update

\begin{equation}
J_{\mathrm{on}}(\theta)
=
\E_{x\sim\ptheta}\!\Big[
\textstyle\sum_t \sg\!\big(A_t(\theta)\big)
\nabla_\theta\log\ptheta(y_t\mid s_t)
\Big].
\label{eq:online}
\end{equation}

Here, $y_t$ denotes the generated token at context $s_t$, and $\sg(\cdot)$ denotes stop-gradient.
The sum includes only model-generated tokens.
This expression uses token summation; the trajectory-wise normalization and advantage clipping used in training are described in Appendix~\ref{app:impl}.
The following proposition characterizes the expected token-level update at a fixed context.

\begin{proposition}[Online update as a divergence gradient]
\label{prop:online}
Fix a context, let $p=\mathrm{softmax}(z)$, and let $q$ have full support.
Then
\begin{enumerate}
\item[(i)]

\begin{equation}
\frac{\partial\KL(p\|q)}{\partial z_i}
=p_i\big[\log(p_i/q_i)-\KL(p\|q)\big];
\end{equation}

\item[(ii)]

\begin{equation}
\E_{y\sim p}
\big[A\,\nabla_z\log p_y\big]
=-\nabla_z\KL(p\|q);
\end{equation}

\item[(iii)]
$\nabla_z\KL(p\|q)=0$ if and only if $p=q$.
Thus, the stationary distribution is unique, while its logits are unique only up to a common additive shift.
\end{enumerate}
\end{proposition}

\begin{proof}
Differentiating $\KL(p\|q)$ gives

\[
\nabla_z\KL(p\|q)
=\sum_j(\log p_j-\log q_j+1)\nabla_zp_j.
\]

The contribution of the constant term vanishes because
$\sum_j\nabla_zp_j=0$.
Using $\nabla_zp_j=p_j\nabla_z\log p_j$ yields part (ii), and Lemma~\ref{lem:softmax} gives part (i).

For part (iii), positivity of $p_i$ implies that a zero gradient requires
$\log(p_i/q_i)=\KL(p\|q)=K$
for every $i$.
Summing $p_i=q_i e^K$ over $i$ gives $e^K=1$, hence $p=q$.
The converse follows directly from part (i).
\end{proof}

\begin{remark}[Online and empirical offline supervision]
\label{rem:online}
Fresh sampling constrains the full distribution through the expected update at a fixed context.
By contrast, one fixed recorded token yields the minimizer set characterized in Proposition~\ref{prop:manifold}.
This comparison concerns expected state-wise updates and does not establish convergence of online training with shared parameters or changing context distributions.
\end{remark}

\subsection{Trajectory-Level Divergence}
\label{app:chi2}

\begin{proposition}[Trajectory-level divergence]
\label{prop:chi2}
Let $P_\theta$ and $P_{\mathrm{ref}}$ be the joint distributions of prompts and complete interaction trajectories induced by $\ptheta$ and $\pref$. Assume a common prompt distribution, shared environment dynamics, a finite horizon, and $P_\theta\ll P_{\mathrm{ref}}$. Then

\[
\chi^2(P_\theta\|P_{\mathrm{ref}})\ge\exp\Big(\E_{x\sim P_\theta}\big[\textstyle\sum_t\KL(p_t\|\pi_t)\big]\Big)-1,
\]

where the sum runs over model-generated tokens.
\end{proposition}

\begin{proof}
For $P\ll Q$, Jensen's inequality for the concave logarithm gives $\KL(P\|Q)=\E_P[\log\frac{dP}{dQ}]\le\log\E_P[\frac{dP}{dQ}]=\log\E_Q[(\frac{dP}{dQ})^2]=\log\big(1+\chi^2(P\|Q)\big)$.
Under the stated assumptions, the prompt and environment factors cancel in the likelihood ratio, so $\log\frac{dP_\theta}{dP_{\mathrm{ref}}}(x)=\sum_t\log\frac{p_t(a_t)}{\pi_t(a_t)}$ over model-generated tokens.
Taking the expectation under $P_\theta$ and conditioning on each context gives $\KL(P_\theta\|P_{\mathrm{ref}})=\E_{x\sim P_\theta}[\sum_t\KL(p_t\|\pi_t)]$.
Combining the two relations and exponentiating yields the claim.
\end{proof}

\subsection{Construction of the Cached Target}
\label{app:trustregion}

At a selected position, let $S\subsetneq\mathcal V$ be the nonempty candidate set and $B=\mathcal V\setminus S$ the nonempty residual bucket.
For any distribution $r$, write $r_B=\sum_{k\in B}r_k$.
Its coarsened distribution consists of the candidate probabilities $(r_j)_{j\in S}$ and the bucket probability $r_B$.
In this subsection, $j$ indexes the cells in $S\cup\{B\}$, and
$\delta_j=\log q_j-\log\pi_j$.
All cell probabilities of $q$ and $\pi$ are positive.

\begin{proposition}[KL-regularized target construction]
\label{prop:trustregion}
Let $\Delta$ denote the simplex over the cached cells.
For $\beta\in[0,1]$, the problem

\[
\max_{r\in\Delta}
\big[\beta\,\E_r[\delta]-\KL(r\|\pi)\big]
\]

has the unique full-support solution

\begin{equation}
\begin{aligned}
\hq_{\beta,j}
&=\frac{\pi_j e^{\beta\delta_j}}{Z_\beta}
=\frac{\pi_j^{1-\beta}q_j^\beta}{Z_\beta},\\
Z_\beta
&=\sum_{k\in S\cup\{B\}}\pi_k^{1-\beta}q_k^\beta.
\end{aligned}
\label{eq:app-hq}
\end{equation}

In particular, $\hq_0=\pi$ and $\hq_1=q$ on the cached cells.
\end{proposition}

\begin{proof}
By Equation~\eqref{eq:app-hq},

\[
\log\hq_{\beta,j}
=\log\pi_j+\beta\delta_j-\log Z_\beta.
\]

Substituting into the objective gives

\[
\beta\,\E_r[\delta]-\KL(r\|\pi)
=\log Z_\beta-\KL(r\|\hq_\beta).
\]

The right-hand side is maximized uniquely at $r=\hq_\beta$.
Positivity follows from the full support of $\pi$ and $q$ on the cells.
\end{proof}

\begin{remark}[Order of aggregation and interpolation]
The target is constructed by first aggregating the reference and teacher distributions into cached cells and then interpolating those cell distributions.
This generally differs from interpolating full-vocabulary distributions and subsequently aggregating them, since

\[
\pi_B^{1-\beta}q_B^\beta
\neq
\sum_{k\in B}\pi_k^{1-\beta}q_k^\beta
\]

in general.
All target-matching statements below concern the cached-cell target in Equation~\eqref{eq:app-hq}.
\end{remark}

\begin{proposition}[$\beta$ bounds the displacement of the target]
\label{prop:betabound}
For $\beta\in[0,1]$,

\begin{equation}
\KL(\pi\|\hq_\beta)
\le\beta\,\KL(\pi\|q),
\label{eq:app-beta-bound}
\end{equation}

where all distributions are defined on the cached cells.
Equality holds if and only if $\beta\in\{0,1\}$ or $q=\pi$ on those cells.
\end{proposition}

\begin{proof}
Equation~\eqref{eq:app-hq} gives

\[
\KL(\pi\|\hq_\beta)
=\beta\,\KL(\pi\|q)+\log Z_\beta.
\]

For $0<\beta<1$, Jensen's inequality yields

\[
Z_\beta
=\E_\pi[(q_j/\pi_j)^\beta]
\le\big(\E_\pi[q_j/\pi_j]\big)^\beta
=1.
\]

Strict concavity implies equality only when $q_j/\pi_j$ is constant across cells, which requires $q=\pi$.
The endpoint cases follow directly from $\hq_0=\pi$ and $\hq_1=q$.
\end{proof}

\begin{remark}[Interpretation of $\beta$]
\label{rem:beta}
For $\beta>0$, dividing the variational objective by $\beta$ makes $\beta^{-1}$ the coefficient of the KL penalty toward the reference.
Equation~\eqref{eq:app-beta-bound} also bounds the target's displacement in the direction $\KL(\pi\|\hq_\beta)$.
These properties concern target construction, not the size of each student update or a constraint on the optimization trajectory.
At $\beta=0$, the target equals the reference on selected positions, while unselected positions still receive scalar-advantage supervision.
\end{remark}

\subsection{Distributional Gradients and Cell-wise Minimizers}
\label{app:gradient}

Write $\hq=\hq_\beta$ for the fixed cached-cell target.
The per-position loss is

\begin{equation}
\ell(z)
=\sum_{j\in S}\hq_j\log\frac{\hq_j}{p_j}
+\hq_B\log\frac{\hq_B}{p_B},
\label{eq:app-ell}
\end{equation}

where $p_B=\sum_{k\in B}p_k$ is recomputed from the current student.

\begin{proposition}[Gradient of the distributional loss]
\label{prop:gradient}
For every finite $z$,

\begin{equation}
\frac{\partial\ell}{\partial z_i}
=
\begin{cases}
p_i-\hq_i, & i\in S,\\[2pt]
p_i\big(1-\hq_B/p_B\big), & i\in B.
\end{cases}
\end{equation}

\end{proposition}

\begin{proof}
Up to a constant independent of $z$,

\[
\ell(z)
=-\sum_{j\in S}\hq_j\log p_j-\hq_B\log p_B.
\]

For $i\in S$, Lemma~\ref{lem:softmax} and Corollary~\ref{cor:group} give

\[
\frac{\partial\ell}{\partial z_i}
=-\hq_i+p_i(1-\hq_B)+\hq_Bp_i
=p_i-\hq_i.
\]

For $i\in B$, the same identities give

\[
\frac{\partial\ell}{\partial z_i}
=p_i(1-\hq_B)
-\hq_B\left(\frac{p_i}{p_B}-p_i\right)
=p_i\left(1-\frac{\hq_B}{p_B}\right).
\]

\end{proof}

\begin{remark}[Candidate-specific supervision]
\label{rem:gradient}
The scalar-regression descent direction is $A(e_y-p)$, whose component for an alternative token $i\neq y$ is $-Ap_i$.
It therefore does not encode separate teacher targets for those alternatives.
For the distributional loss, the descent direction for each candidate logit is $\hq_i-p_i$, providing a target-specific correction.
Within the bucket, all logit gradients share the factor $1-\hq_B/p_B$, which vanishes when the total bucket mass is matched.
\end{remark}

\label{app:unique}
\begin{proposition}[Cell-wise minimizers and logit rescaling]
\label{prop:unique}
Assume $\hq_j>0$ for every cell $j\in S\cup\{B\}$.
Then:
\begin{enumerate}
\item[(a)]
$\nabla_z\ell(z)=0$ if and only if $p_j=\hq_j$ for every $j\in S$ and $p_B=\hq_B$.
The minimizer on the coarsened simplex is unique, while the full-vocabulary minimizers retain $|B|-1$ degrees of freedom in probability space.
\item[(b)]
Let $z^\star$ be a minimizer and $\alpha>0$.
If $\hq$ is not uniform on $S$, or if
$\hq_B<\min_{j\in S}\hq_j$,
then $\alpha z^\star$ is a minimizer only when $\alpha=1$.
\item[(c)]
Let $z_0$ be reference logits with $\mathrm{softmax}(z_0)=\pi$.
For $\alpha>0$, $\alpha z_0$ is a minimizer if and only if

\begin{equation}
\frac{\pi_j^\alpha}{\sum_{k\in\mathcal V}\pi_k^\alpha}
=\hq_j
\qquad\text{for every }j\in S.
\label{eq:app-ray-matching}
\end{equation}

A necessary condition is therefore

\begin{equation}
\frac{\hq_j}{\hq_k}
=\left(\frac{\pi_j}{\pi_k}\right)^\alpha
\qquad(j,k\in S).
\label{eq:app-ray}
\end{equation}

For the interpolated target, this condition becomes

\begin{equation}
\left(\frac{q_j}{q_k}\right)^\beta
=
\left(\frac{\pi_j}{\pi_k}\right)^{\alpha-1+\beta}.
\label{eq:app-ray-interpolation}
\end{equation}

\end{enumerate}
\end{proposition}

\begin{proof}
For part (a), Proposition~\ref{prop:gradient} shows that a zero gradient on $S$ requires $p_j=\hq_j$ for each candidate.
Normalization then gives $p_B=\hq_B$, which also makes every bucket-logit gradient zero.
Conversely, matching these cell probabilities makes the full gradient zero.
The KL divergence on the coarsened simplex attains its unique minimum of zero at the target.
The individual probabilities within $B$ remain positive and sum to $\hq_B$, leaving $|B|-1$ degrees of freedom.

For part (b), if $z^\star$ and $\alpha z^\star$ both minimize the loss, then for all $i,j\in S$,

\[
\frac{\hq_i}{\hq_j}
=e^{z_i^\star-z_j^\star}
=e^{\alpha(z_i^\star-z_j^\star)}.
\]

If the candidate targets are not all equal, this requires $\alpha=1$.
Under the alternative condition
$\hq_B<\min_{j\in S}\hq_j$,
every bucket probability is smaller than every candidate probability, so every bucket logit is smaller than every candidate logit.
Consequently,

\[
R(\alpha)
=\log\sum_{j\in S}e^{\alpha z_j^\star}
-\log\sum_{k\in B}e^{\alpha z_k^\star}
\]

is strictly increasing.
Matching the target requires
$R(\alpha)=\log((1-\hq_B)/\hq_B)$,
which already holds at $\alpha=1$ and cannot hold elsewhere.

For part (c), rescaling the reference logits gives

\[
p_j(\alpha z_0)
=\frac{\pi_j^\alpha}{\sum_{k\in\mathcal V}\pi_k^\alpha}.
\]

Part (a) therefore gives Equation~\eqref{eq:app-ray-matching}.
Taking candidate ratios yields Equation~\eqref{eq:app-ray}.
Substituting
$\hq_j\propto\pi_j^{1-\beta}q_j^\beta$
gives Equation~\eqref{eq:app-ray-interpolation}.
\end{proof}

\begin{remark}[Scope of the rescaling characterization]
\label{rem:unique}
Part (b) concerns rescaling logits that already minimize the distributional loss, whereas part (c) concerns reaching a minimizer from the reference.
The ratio condition in Equation~\eqref{eq:app-ray} is necessary but not sufficient: the absolute candidate probabilities, and hence the total bucket mass, must also match.
Thus, a rescaling that satisfies the scalar objective need not minimize the distributional objective.
However, rescaling can still be optimal when the cached target satisfies Equation~\eqref{eq:app-ray-matching}.
The result therefore does not exclude every logit contraction or every increase in entropy.
\end{remark}

\subsection{The Role of the Bucket}
\label{app:bucket}

\begin{proposition}[Role of the bucket]
\label{prop:bucket}
The following statements hold.
\begin{enumerate}
\item[(a)]
Let $\hq_{\mathrm{full}}$ be any full-vocabulary distribution whose coarsening equals the cached target $\hq$.
Write $r_{\cdot\mid B}$ for the conditional distribution of $r$ within $B$.
Then

\begin{equation}
\begin{split}
\KL(\hq_{\mathrm{full}}\|p)
&=\ell(z)\\
&\quad+\hq_B
\KL(\hq_{\mathrm{full},\cdot\mid B}\|p_{\cdot\mid B}).
\end{split}
\label{eq:app-chainrule}
\end{equation}

The remainder satisfies

\begin{equation}
\begin{split}
0
&\le\hq_B
\KL(\hq_{\mathrm{full},\cdot\mid B}\|p_{\cdot\mid B})\\
&\le\hq_B\log\frac1{\min_{i\in B}p_{i\mid B}}.
\end{split}
\label{eq:app-remainder}
\end{equation}

\item[(b)]
If the bucket term is removed and both target and student distributions are renormalized on $S$, the resulting KL loss depends only on logit differences within $S$.
It is invariant under
$z\mapsto z+s\sum_{k\in B}e_k$,
along which $p_B$ ranges over $(0,1)$.
\end{enumerate}
\end{proposition}

\begin{proof}
For part (a), factor the bucket probabilities as
$\hq_{\mathrm{full},k}=\hq_B\hq_{\mathrm{full},k\mid B}$
and $p_k=p_Bp_{k\mid B}$.
The bucket contribution to the full KL separates into
$\hq_B\log(\hq_B/p_B)$
and the conditional KL term in Equation~\eqref{eq:app-chainrule}.
The lower bound follows from nonnegativity of KL.
For conditional distributions $u,v$,

\[
\begin{aligned}
\KL(u\|v)
&=\sum_i u_i\log u_i+\sum_i u_i\log(1/v_i)\\
&\le\log\frac1{\min_i v_i},
\end{aligned}
\]

which gives the upper bound.

For part (b), the renormalized student probability of $j\in S$ is
$e^{z_j}/\sum_{k\in S}e^{z_k}$,
which is independent of every bucket logit.
Let
$u=\sum_{j\in S}e^{z_j}>0$
and
$w=\sum_{k\in B}e^{z_k}>0$.
After the indicated shift,

\[
p_B=\frac{e^s w}{u+e^s w},
\]

which ranges over $(0,1)$ as $s$ varies over $\R$.
\end{proof}

\begin{remark}[What bucket supervision preserves]
\label{rem:bucket}
The bucket constrains the total probability outside the candidate set while leaving its internal allocation unspecified.
The omitted conditional KL is weighted by $\hq_B$, but a small bucket mass alone does not guarantee a small remainder.
The full-vocabulary extension in part (a) need not equal the target obtained by interpolating full-vocabulary reference and teacher distributions.
Part (b) specifically concerns losses that renormalize both distributions on $S$.
\end{remark}

\subsection{The Single-Coordinate Case}
\label{app:projection}

\begin{proposition}[One coordinate]
\label{prop:projection}
Take $S=\{y\}$ and $\beta=1$, so that
$\hq_y=q_y$ and $\hq_B=1-q_y$.
Then

\begin{equation}
\ell(z)
=\hq_y\log\frac{\hq_y}{p_y}
+(1-\hq_y)\log\frac{1-\hq_y}{1-p_y},
\end{equation}

and

\begin{equation}
\begin{split}
\nabla_z\ell
&=\frac{p_y-\hq_y}{1-p_y}(e_y-p),\\
\nabla_zL_{\mathrm{pos}}
&=(\log p_y-\log\hq_y)(e_y-p).
\end{split}
\label{eq:app-binary-grad}
\end{equation}

Away from $p_y=\hq_y$, the gradients are positive scalar multiples of each other.
Both losses have the same minimizer set $M_c$, with $c=\log q_y$.
\end{proposition}

\begin{proof}
Proposition~\ref{prop:gradient} gives the $y$-component
$p_y-\hq_y$.
For $i\neq y$, it gives

\[
p_i\left(1-\frac{1-\hq_y}{1-p_y}\right)
=-\frac{p_i(p_y-\hq_y)}{1-p_y}.
\]

These components yield the first identity in Equation~\eqref{eq:app-binary-grad}.
The second follows from Proposition~\ref{prop:manifold}.
Because the logarithm is strictly increasing, both scalar coefficients have the sign of $p_y-\hq_y$ and vanish exactly when $p_y=\hq_y$.
\end{proof}

\begin{remark}[Supervision coverage and loss choice]
\label{rem:projection}
With only the recorded token and a bucket, forward KL does not remove the underdetermination of alternative-token probabilities.
At $\beta=1$, it shares the scalar-regression minimizers, although its gradient magnitudes differ.
This equivalence is state-wise and does not imply identical updates after aggregating multiple positions through shared parameters.
For $\beta<1$, the two-cell distributional objective instead matches the interpolated target and still leaves the within-bucket allocation unspecified.
These observations distinguish the information retained by the supervision from the choice of loss used to fit it.
\end{remark}

\section{Experimental Details}
\label{app:impl}

This appendix describes the datasets, training configurations, cache construction, evaluation protocols, and computational costs of our experiments.
We also specify the configurations used for the hyperparameter analysis, the comparison of KL directions, and the distributional diagnostics.

\subsection{Models and Reference Initialization}
\label{app:impl-models}

We use Qwen3-4B-Base \citep{qwen3} as the student and consider MiroThinker-1.7-mini \citep{mirothinker} and Qwen3-32B \citep{qwen3} as teachers.
To keep the teacher consistency proposed in \lopd{} \citep{wu2026lightning}, for each teacher, we construct a separate reference policy $\pref$ by supervised fine-tuning on trajectories generated by that teacher.
The same teacher subsequently scores the reference rollouts, satisfying the teacher-consistency condition of \lopd{}.
Within each teacher setting, all distillation methods are initialized from the same reference checkpoint.

For agentic search, both teachers generate trajectories for the same set of SFT questions in the same retrieval environment.

\paragraph{Filtering policy.}
We retain completed trajectories regardless of answer correctness.
Failed or incomplete generations are retried up to two additional times, and questions without a completed trajectory after three attempts are excluded from SFT.
No trajectory reached the eight-turn limit during SFT data generation.

\paragraph{Model checkpoints.}
We initialize the student backbone from the official \texttt{Qwen/Qwen3-4B-Base} checkpoint and use \texttt{Qwen/Qwen3-32B} and \texttt{miromind-ai/MiroThinker-1.7-mini} as teachers.
For each teacher, we use the final SFT checkpoint as $\pref$ and initialize all distillation methods from this checkpoint.
We evaluate the final checkpoint of each distillation run.

\paragraph{Vocabulary alignment.}
The student and both teachers share the same vocabulary and token-ID mapping.
This allows recorded tokens and candidate alternatives to be matched directly across the reference, student, and teacher distributions.
No vocabulary remapping is required for teacher scoring or target construction.

\subsection{Agentic Search}
\label{app:impl-search}

\paragraph{Datasets.}
Table~\ref{tab:data} summarizes the question sources and counts for SFT, distillation, and evaluation.
The SFT questions are drawn from the training splits of Natural Questions, HotpotQA, and MuSiQue.
For distillation, we select $5{,}000$ questions each from the training splits of HotpotQA and MuSiQue and sample four trajectories per question from the frozen reference policy.
We evaluate on the development splits of HotpotQA, 2WikiMultiHop, and MuSiQue, together with Bamboogle and the FRAMES test split.
The question-subset selection procedure and random seed is $42$.

\begin{table}[t]
\centering
\footnotesize
\setlength{\tabcolsep}{3pt}
\begin{tabular}{@{}llr@{}}
\toprule
\tableheader{Stage} & \tableheader{Source} & \tableheader{Questions} \\
\midrule
SFT
& Natural Questions (train) & $1{,}200$ \\
& HotpotQA (train) & $2{,}000$ \\
& MuSiQue (train) & $800$ \\
\midrule
Distillation
& HotpotQA (train) & $5{,}000$ \\
& MuSiQue (train) & $5{,}000$ \\
\midrule
Evaluation
& HotpotQA (dev) & $1{,}000$ \\
& 2WikiMultiHop (dev) & $1{,}000$ \\
& MuSiQue (dev) & $1{,}000$ \\
& Bamboogle & $125$ \\
& FRAMES (test) & $824$ \\
\bottomrule
\end{tabular}
\caption{Question counts for agentic search.
For each teacher setting, we generate four reference-policy rollouts per distillation question and retain $38{,}400$ trajectories after filtering and random subsampling.}
\label{tab:data}
\end{table}

\paragraph{Retrieval environment.}
We use the Search-R1 retrieval service \citep{searchr1} with its publicly released Wikipedia 2018 corpus \citep{wiki18dpr} and precomputed FAISS index.
We retain the released passage segmentation and use \texttt{intfloat/e5-base-v2} \citep{e5} as the retriever.
Each search returns the top three passages, which are appended to the trajectory as observations.
Retrieved tokens remain available as context but are excluded from the training loss.

We implement rollout generation and interaction control separately, with an eight-turn interaction budget during training and evaluation.
At the turn limit, the environment inserts an answer prefix as a masked observation and requests a final completion.

\paragraph{Reference rollouts and online sampling.}
For each teacher setting, we sample four trajectories per question from the frozen reference policy $\pref$ before optimization.
Using $5{,}000$ questions each from HotpotQA and MuSiQue yields $40{,}000$ trajectories before filtering.
After filtering, we randomly subsample the remaining trajectories to retain $38{,}400$ trajectories in each teacher setting.
Within each teacher setting, \lopd{} and \shortours{} use exactly the same retained trajectories.
We randomly shuffle the cached trajectories and train for one pass without replacement.
Online OPD uses the same training-question pool and samples four trajectories per question from the current student, with teacher scores computed on the newly generated trajectories.

\begin{table}[t]
\centering
\footnotesize
\setlength{\tabcolsep}{3pt}
\begin{tabular}{@{}lccc@{}}
\toprule
\tableheader{Stage} & \tableheader{Temp.} & \tableheader{Top-$p$} & \tableheader{Responses/question} \\
\midrule
Teacher SFT generation & $0.8$ & $0.95$ & $1$ \\
Reference rollouts & $0.8$ & $0.95$ & $4$ \\
Online OPD rollouts & $0.8$ & $0.95$ & $4$ \\
Evaluation & $0.8$ & $0.95$ & $4$ \\
\bottomrule
\end{tabular}
\caption{Generation settings for agentic search.
Responses/question denotes the number of trajectories sampled per question at each stage.}
\label{tab:rollout-settings}
\end{table}

\paragraph{Optimization.}
All three OPD methods use the optimization settings in Table~\ref{tab:hparams}.
Training consists of $150$ optimizer steps with a global batch size of $256$ trajectories.
For the offline methods, this corresponds to one complete pass over the $38{,}400$ cached trajectories.
We average the loss over scored tokens within each trajectory and then average over trajectories in the global batch, giving each trajectory equal weight.
Trajectories are processed individually, with gradients accumulated in FP32 to form the global batch.
For \shortours{}, selected and unselected positions contribute to the same within-trajectory average, without separate normalization or additional weighting.
Table~\ref{tab:methods} summarizes the method-specific supervision and update rules.

A shared scoring mask is used for student optimization, online teacher scoring, and offline score caching.
Scored positions include model-generated reasoning, search queries, final answers, their format tokens, and the terminal \texttt{<|im\_end|>} token.
Prompts, retrieved passages, environment-generated messages, and the environment-prefilled answer prefix are excluded from the loss but remain available as context.
Model-generated tokens following the prefilled prefix are scored normally.
The initial token position is never scored, and no padding is introduced because trajectories are processed individually.

The main results use one training run per configuration with seed $0$ for reference-rollout sampling and training-data ordering.
We evaluate the final checkpoint of each run.
The four sampled responses per evaluation question are inference samples rather than independent training runs.
We use AdamW with a constant learning rate of $2\times10^{-6}$, weight decay $0.1$, and momentum coefficients $(0.9,0.98)$.
No importance weighting, PPO clipping, advantage normalization, auxiliary KL regularization, or gradient clipping is applied.

\begin{table}[!tp]
\centering
\footnotesize
\setlength{\tabcolsep}{3pt}
\begin{tabularx}{\columnwidth}{@{}lX@{}}
\toprule
\tableheader{Setting} & \tableheader{Value} \\
\midrule
\multicolumn{2}{@{}l}{\tableheader{\textit{Environment}}} \\
Passages per search & $3$ \\
Turn cap & $8$ \\
Max.\ sequence length & $30{,}208$ tokens \\
Max.\ generated tokens per turn & $2{,}560$ \\
Max.\ retrieved tokens per turn & No separate limit \\
Sampling precision & bf16 \\
\midrule
\multicolumn{2}{@{}l}{\tableheader{\textit{Distillation}}} \\
Optimizer & AdamW \\
Adam coefficients & $(0.9,0.98)$ \\
Weight decay & $0.1$ \\
Learning rate & $2\times10^{-6}$ \\
Learning-rate schedule & Constant \\
Warmup & None \\
Gradient clipping & None \\
Optimizer steps & $150$ \\
Global batch size & $256$ trajectories \\
Microbatch size & $1$ trajectory \\
Gradient accumulation & $256$ trajectories per optimizer update \\
Parameter and gradient precision & fp32 \\
Forward-pass precision & bf16 \\
Trainable parameters & Full model \\
Gradient checkpointing & Enabled \\
Loss masking & Model-generated tokens only \\
\midrule
\multicolumn{2}{@{}l}{\tableheader{\textit{SFT}}} \\
Optimizer & AdamW \\
Adam coefficients & $(0.9,0.999)$ \\
Weight decay & $0$ \\
Optimizer steps & $3{,}000$ \\
Learning rate & $8\times10^{-5}$ \\
Learning-rate schedule & Cosine decay \\
Warmup ratio & $0.1$ \\
Global batch size & $256$ \\
Max.\ sequence length & $16{,}384$ tokens \\
Sequence packing & Enabled \\
Precision & bf16 \\
Trainable parameters & Full model \\
Loss masking & Model-generated tokens only \\
\bottomrule
\end{tabularx}
\caption{Environment and optimization settings for agentic search.
The environment sequence-length limit includes prompt tokens, retrieved passages, environment-inserted text, and model-generated tokens.
SFT follows the 4B training configuration of \citet{wu2026lightning}, with the interaction-aware loss mask described in the text.
Hardware allocation is reported in Appendix~\ref{app:cost}.}
\label{tab:hparams}
\end{table}

\begin{table*}[!tp]
\centering
\footnotesize
\setlength{\tabcolsep}{4pt}
\begin{tabularx}{\textwidth}{@{}p{0.20\textwidth}XXX@{}}
\toprule
\tableheader{Setting} & \tableheader{Online OPD} & \tableheader{\lopd{}} & \tableheader{\shortours{}} \\
\midrule
Rollout source
& Current student
& Frozen reference
& Same fixed reference rollouts as \lopd{} \\
Teacher inference during optimization
& Required
& Not required
& Not required \\
Supervision
& Teacher scores on fresh student rollouts
& Cached teacher scores for recorded tokens
& Fixed targets constructed from cached reference and teacher scores \\
Update
& Sampled-token policy-gradient update
& Equation~\eqref{eq:offline}
& Equation~\eqref{eq:dist-loss} at all scored positions \\
Clipping
& Current token-level advantage clipped to $[-10,10]$
& Current token-level advantage clipped to $[-10,10]$
& Fixed teacher--reference gap clipped to $[-10,10]$ for two-cell target construction \\
Position selection
& --
& --
& $x_{2,t}>-8$ \\
Selected-position partition
& --
& --
& Candidate tokens and a bucket \\
Unselected-position partition
& --
& --
& Recorded token and a bucket \\
\bottomrule
\end{tabularx}
\caption{Method-specific configurations.
Here, $x_{2,t}$ denotes the second-highest reference token log-probability at position $t$.
For \shortours{}, position selection determines the candidate set, while both selected and unselected positions use forward KL with fixed targets.
Shared optimization settings are listed in Table~\ref{tab:hparams}.}
\label{tab:methods}
\end{table*}

\subsection{Cache Construction and Loss Computation}
\label{app:impl-cache}

\paragraph{Cached supervision.}
Both offline methods cache the teacher log-probability of each recorded token.
For \shortours{}, we additionally retain reference scores for recorded tokens and cache reference and teacher scores over $\Sset_t$ at selected positions.
Reference and teacher scores are computed using Transformers and vLLM, respectively, from the original model distributions without sampling-time temperature scaling or truncation.
Teacher scores for candidate tokens are retrieved from the returned top-$256$ entries and the recorded-token score; unavailable candidate probabilities are approximated as zero, making the resulting teacher bucket mass an upper bound.
Targets are reconstructed from the cached scores and remain fixed throughout training.
Table~\ref{tab:cache} summarizes the cached supervision.

\begin{table}[t]
\centering
\footnotesize
\setlength{\tabcolsep}{2.5pt}
\resizebox{\columnwidth}{!}{%
\begin{tabular}{@{}lccc@{}}
\toprule
& \tableheader{\lopd{}} & \multicolumn{2}{c}{\tableheader{\shortours{}}} \\
\cmidrule(lr){3-4}
\tableheader{Required item} & & \tableheader{Selected} & \tableheader{Unselected} \\
\midrule
Trajectory tokens and loss mask
& \checkmark & \checkmark & \checkmark \\
Recorded-token teacher log-probability
& \checkmark & \checkmark & \checkmark \\
Recorded-token reference log-probability
& -- & \checkmark & \checkmark \\
Candidate token IDs
& -- & \checkmark & -- \\
Reference candidate log-probabilities
& -- & \checkmark & -- \\
Teacher candidate log-probabilities
& -- & \checkmark & -- \\
Reference bucket mass
& -- & Derived & Derived \\
Teacher bucket mass
& -- & \checkmark & -- \\
\bottomrule
\end{tabular}%
}
\caption{Information used to construct offline supervision.
Derived bucket masses are computed from reference probabilities.
At selected positions, unavailable teacher candidate probabilities are approximated as zero, so the teacher bucket mass is an upper bound on the true remaining mass.
The recorded token also belongs to the selected candidate set.}
\label{tab:cache}
\end{table}

\paragraph{Storage and numerical computation.}
Trajectory caches are stored as uncompressed \texttt{.pt} files with int64 token IDs and fp32 teacher log-probabilities.
Reference and teacher candidate caches use compressed \texttt{.npz} files with int32 token IDs and fp32 scores.
Cache merging and reference and teacher residual-mass calculations use float64, after which the resulting log-probabilities are converted to fp32.
Student log-probabilities, target normalization, and KL evaluation are computed in fp32.
Residual bucket masses are lower-bounded by $10^{-12}$ before taking logarithms.
Invalid padded entries use a finite log-space sentinel of $-10^{30}$ and are masked out of the KL sum.
Targets are reconstructed from immutable cached scores and remain fixed throughout optimization.

\paragraph{Training loss.}
During training, student probabilities are recomputed at each recorded context using full-vocabulary normalization.
The student bucket mass is computed as one minus the summed candidate probabilities, with the numerical safeguards described above.
All scored positions use the forward-KL objective in Equation~\eqref{eq:dist-loss}.
Selected positions use the candidate-token and bucket partition, while unselected positions use the two-cell partition consisting of the recorded token and the bucket.
At unselected positions, clipping applies to the fixed teacher--reference gap used to construct the target, rather than to a recomputed teacher--student advantage.
Losses are averaged over scored positions within each trajectory and then over trajectories in the global batch.
Selected and unselected positions contribute to the same trajectory-level mean without separate weighting.
All supervision is obtained from the fixed caches, with no teacher inference during optimization.

\subsection{Evaluation}
\label{app:eval}

We evaluate the SFT reference, all three distillation methods, and the teachers using the same agentic-search protocol.
For each question, we sample four responses and average their individual EM and token-level F1 scores, without majority voting or best-response selection.
Benchmark scores are obtained by averaging these question-level scores.

We extract the text between the first \texttt{<answer>} tag and the first subsequent \texttt{</answer>} tag in the final action.
For forced final completions, the environment-provided opening tag is restored before extraction.
Answers are normalized by lowercasing, removing punctuation and articles, and collapsing whitespace.
EM and token-level F1 are computed separately against each accepted gold answer, retaining the highest value for each metric.
Empty answers and generations without a complete answer span receive zero for both metrics and remain included in the averages.
No fallback extraction is applied to untagged text.

We use a custom evaluation implementation at repository commit \texttt{745fb20}, with SQuAD-style answer normalization and token-level F1.
Retrieval requests are retried up to four times after transport errors.
Unrecoverable retrieval or inference errors interrupt evaluation rather than producing zero-scored samples.
Completed questions are saved, and interrupted runs resume from the remaining questions, yielding four scored responses per question.

\begin{table}[t]
\centering
\footnotesize
\setlength{\tabcolsep}{3pt}
\begin{tabularx}{\columnwidth}{@{}lX@{}}
\toprule
\tableheader{Setting} & \tableheader{Value} \\
\midrule
Responses per question & $4$ \\
Response aggregation & Mean of individual scores \\
Temperature & $0.8$ \\
Top-$p$ & $0.95$ \\
Top-$k$ sampling & Disabled \\
Turn cap & $8$ \\
Max.\ sequence length & $30{,}208$ tokens \\
Max.\ generated tokens per turn & $2{,}560$ \\
Metrics & EM and token-level F1 ($\times100$) \\
Gold-alias handling & Maximum over aliases for each metric \\
Incomplete-output handling & Zero scores; included in averages \\
Sampling seed & $1000i+s$, $s\in\{0,1,2,3\}$ \\
Checkpoint selection & Final checkpoint \\
\bottomrule
\end{tabularx}
\caption{Evaluation protocol for agentic search.
Here, $i$ denotes the question index and $s$ the response index.
All methods use the same sampling seeds, interaction budgets, and scoring procedure across the five benchmarks.
Forced final completions use the same per-turn generation limit.}
\label{tab:eval}
\end{table}

\subsection{Hyperparameter and Loss Comparisons}
\label{app:impl-ablation}

We evaluate the effects of $\beta$, $\rho$, and the KL direction on 2WikiMultiHop and FRAMES.
The $\beta$ and $\rho$ studies include both teachers, while the KL-direction comparison uses Qwen3-32B.
Table~\ref{tab:variants} lists the changes from the default configuration.
All variants use training seed $0$.
Within each comparison, all variants use the same reference checkpoint, training rollouts, evaluation questions, and decoding settings.

\paragraph{Interpolation coefficient.}
We vary $\beta$ while holding the candidate sets, cached scores, and remaining training settings fixed.
For each value of $\beta$, we reconstruct the targets at both selected and unselected positions using the definitions in Section~\ref{sec:method-loss}.
The targets remain fixed throughout each training run.

\paragraph{Teacher-selection threshold.}
We vary $\rho$ and $\beta=0.5$.
For each value of $\rho$, we reconstruct the candidate sets, bucket masses, and targets from the same cached scoring outputs without additional teacher inference.
Reusing these outputs ensures that all variants use the same underlying scores.
Because teacher selection is capped and combined with reference candidates, changing $\rho$ need not change the final candidate set at every position.

\paragraph{KL direction.}
The reverse-KL variant replaces forward KL at every scored position with

\begin{equation}
\ell_t^{\mathrm{rev}}(\theta)
=
\sum_{c\in\mathcal C_t}
\ptheta(c\mid s_t)
\log\frac{\ptheta(c\mid s_t)}{\hq_{\beta,t}(c)}.
\label{eq:app-reverse-kl}
\end{equation}

This replacement applies to both the candidate-token partitions at selected positions and the two-cell partitions at unselected positions.
Both variants use the same cached scores, candidate sets, fixed targets, and loss-reduction scheme.
The reverse-KL variant uses the same numerical precision and safeguards described in Appendix~\ref{app:impl-cache}, without additional stabilization.

\begin{table}[t]
\centering
\footnotesize
\setlength{\tabcolsep}{3pt}
\begin{tabular}{@{}lp{0.61\columnwidth}@{}}
\toprule
\tableheader{Experiment} & \tableheader{Change from default \shortours{}} \\
\midrule
$\beta$ sweep
& $\beta\in\{0,0.25,0.5,0.75,1\}$; $\rho=0.99$ \\
$\rho$ sweep
& $\rho\in\{0.5,0.8,0.9,0.95,0.99\}$; $\beta=0.5$ \\
KL direction
& Forward or reverse KL on the same cached cells and targets \\
\bottomrule
\end{tabular}
\caption{Configurations for the analyses in Section~\ref{sec:further}. All variants use training seed $0$. Within each comparison, the reference initialization, training rollouts, optimization budget, and evaluation protocol remain unchanged.}
\label{tab:variants}
\end{table}

\subsection{Mathematical Reasoning}
\label{app:impl-math}

We follow the 4B training recipe of \citet{wu2026lightning}, using Qwen3-4B-Base as the student and Qwen3-32B as the teacher.
We construct the SFT data from OpenThoughts-3 prompts with responses generated by Qwen3-32B, ensuring that the same teacher is used for SFT data generation and distillation.
SFT runs for $3{,}000$ optimizer steps with a global batch size of $256$, a maximum sequence length of $16{,}384$ tokens, and sequence packing enabled.
The learning rate is $8\times10^{-5}$ with cosine decay and a warmup ratio of $0.1$.
We use the final SFT checkpoint as $\pref$.
Table~\ref{tab:math-setup} summarizes the configuration.

Distillation uses DAPO-Math-17k.
For the offline methods, we sample one response per training prompt from $\pref$ and cache teacher supervision before optimization.
\lopd{} and \shortours{} share the same recorded trajectories, while online OPD generates fresh responses from the current student.
Training runs for $150$ optimizer steps with a global batch size of $256$ and a constant learning rate of $2\times10^{-6}$.
Training responses are generated with temperature $0.8$, top-$p$ $1.0$, and a maximum response length of $4{,}096$ tokens.
For \shortours{}, candidate selection and target construction follow Section~\ref{sec:method-loss}, with $\beta=0.5$ and $\rho=0.99$.

We evaluate the final checkpoints on AIME 2024 and AIME 2025 using the mathematical-reasoning prompt template of \citet{wu2026lightning}, which requests step-by-step reasoning and a final answer enclosed in \texttt{\textbackslash boxed\{\}}.
For each problem, we sample $32$ solutions with temperature $0.6$, top-$p$ $0.95$, and a maximum generation length of $32{,}768$ tokens.
We estimate pass@1 as the fraction of correct solutions for each problem and average these estimates over problems.

\begin{table}[t]
\centering
\footnotesize
\setlength{\tabcolsep}{3pt}
\begin{tabularx}{\columnwidth}{@{}lX@{}}
\toprule
\tableheader{Setting} & \tableheader{Value} \\
\midrule
Student & Qwen3-4B-Base \\
Teacher & Qwen3-32B \\
\midrule
\multicolumn{2}{@{}l}{\tableheader{\textit{SFT}}} \\
Prompt source & OpenThoughts-3 \\
Response generator & Qwen3-32B \\
Optimizer steps & $3{,}000$ \\
Global batch size & $256$ \\
Learning rate & $8\times10^{-5}$ \\
Learning-rate schedule & Cosine decay \\
Warmup ratio & $0.1$ \\
Max.\ sequence length & $16{,}384$ tokens \\
Sequence packing & Enabled \\
\midrule
\multicolumn{2}{@{}l}{\tableheader{\textit{Distillation}}} \\
Training data & DAPO-Math-17k \\
Responses per offline prompt & $1$ \\
Optimizer steps & $150$ \\
Global batch size & $256$ trajectories \\
Learning rate & $2\times10^{-6}$ \\
Learning-rate schedule & Constant \\
Adam coefficients & $(0.9,0.98)$ \\
Weight decay & $0.1$ \\
Training rollout temperature & $0.8$ \\
Training rollout top-$p$ & $1.0$ \\
Max.\ training response length & $4{,}096$ tokens \\
\shortours{} configuration
& Section~\ref{sec:method-loss};
$\beta=0.5$, $\rho=0.99$ \\
\midrule
\multicolumn{2}{@{}l}{\tableheader{\textit{Evaluation}}} \\
Benchmarks & AIME 2024 and AIME 2025 \\
Solutions per problem & $32$ \\
Temperature & $0.6$ \\
Top-$p$ & $0.95$ \\
Max.\ response length & $32{,}768$ tokens \\
Metric & Average pass@1 ($\times100$) \\
\midrule
Training and evaluation seeds & $0$ \\
Checkpoint selection & Final checkpoint of each stage \\
\bottomrule
\end{tabularx}
\caption{Mathematical-reasoning settings.
We follow the 4B training recipe and mathematical-reasoning evaluation protocol of \citet{wu2026lightning}, using Qwen3-32B for both SFT data generation and distillation.
\shortours{} uses the default configuration in Section~\ref{sec:method-loss}.}
\label{tab:math-setup}
\end{table}

\subsection{Software, Hardware, and Computational Cost}
\label{app:cost}

\paragraph{Software and hardware.}
All experiments use NVIDIA A100 GPUs with $80$\,GB of memory per GPU.
SFT follows the configuration of \citet{wu2026lightning}.
OPD training uses a custom PyTorch trainer with Hugging Face Transformers, PyTorch \texttt{2.11.0+cu130}, and SDPA attention.
Rollout generation, evaluation, and teacher scoring use vLLM \texttt{0.25.1} with tensor-parallel size $1$.
Student optimization runs on a single GPU, with gradient accumulation over $256$ trajectories per optimizer step.
Gradient checkpointing is enabled, and CPU offloading is disabled.

Online OPD uses two GPUs concurrently: one for student optimization, and one for rollout generation and the teacher scoring server.
Both offline methods use a single GPU during optimization and read trajectories and teacher supervision from their caches.
Cache construction is performed before training, and the resulting caches are reused across runs that share the same reference rollouts.
Retrieval runs in a separate CPU service and consumes no GPU resources.

\paragraph{Cache size.}
Table~\ref{tab:cache-size} reports the on-disk cache sizes for the main agentic-search experiments, using the same recorded rollouts within each teacher setting.
The measurements include trajectory token IDs, token-type masks, cached scores, and metadata, including retrieved passage text.
For \shortours{}, the totals additionally include reference and teacher candidate scores and bucket information.
Trajectory files are stored without compression, whereas candidate-score arrays use compressed NumPy archives.
Auxiliary diagnostic and ablation fields stored within these files are included; separate reference-entropy caches are excluded.
Including the shared trajectory cache, \shortours{} occupies $5.40$ and $7.53$\,GiB for MiroThinker-1.7-mini and Qwen3-32B, respectively, compared with $2.16$ and $1.79$\,GiB for \lopd{}.

For comparison, we estimate the storage required for full-vocabulary teacher scores as

\[
\text{size}=N_{\mathrm{gen}}\times V\times b,
\]

where $N_{\mathrm{gen}}$ counts scored model-generated tokens, including the terminal EOS token, $V=151{,}936$ is the vocabulary dimension, and $b=2$ assumes a two-byte representation for each score.
The two caches contain $17{,}947{,}970$ and $27{,}228{,}325$ scored tokens, yielding estimated sizes of $4.96$ and $7.53$\,TiB, respectively.
These estimates assume uncompressed storage and exclude trajectory and metadata overhead.
They describe full-vocabulary teacher supervision rather than the recorded-token scores used by the sampled-token online baseline.

\begin{table}[t]
\centering
\footnotesize
\setlength{\tabcolsep}{3pt}
\begin{tabularx}{\columnwidth}{@{}Xrr@{}}
\toprule
& \tableheader{MiroThinker} & \tableheader{Qwen3-32B} \\
\midrule
Cached trajectories
& $38{,}400$ & $38{,}400$ \\
Model-generated tokens
& $17{,}947{,}970$ & $27{,}228{,}325$ \\
\midrule
\lopd{} cache (GiB)
& $2.16$ & $1.79$ \\
\shortours{} cache (GiB)
& $5.40$ & $7.53$ \\
Relative cache size
& $2.50\times$ & $4.21\times$ \\
Full-vocabulary estimate (TiB)
& $4.96$ & $7.53$ \\
\bottomrule
\end{tabularx}
\caption{Cache statistics for the main agentic-search experiments.
Candidate-token counts are averaged over selected positions and exclude the residual bucket.
Measured cache sizes include shared trajectory storage.
Full-vocabulary estimates assume two bytes per teacher score and exclude trajectory overhead.}
\label{tab:cache-size}
\end{table}

\paragraph{Compute accounting.}
GPU-hours are computed from Slurm \texttt{sacct} elapsed times multiplied by the number of GPUs allocated to each job.
All three methods complete $150$ optimizer steps.
For online OPD, we additionally count teacher-server GPU-hours over the interval overlapping the training job, without apportioning shared server time among concurrent jobs.
Timings include model loading, checkpoint writing, training diagnostics, and retrieval waiting time, but exclude separate evaluation jobs and CPU retrieval-service costs.

Offline preprocessing is reported separately and includes reference-rollout generation with recorded-token scoring, reference candidate scoring, and teacher candidate scoring.
Base-cache construction costs are reported both excluding and including the overlapping shared teacher-server allocation.
Each configuration is measured from a single completed run; the reported cost ranges reflect shared-server accounting rather than variation across repeated runs.
Failed and cancelled attempts are reported separately and included only in the all-allocation total.

\section{Training Diagnostics}
\label{app:diagnosis}

We analyze teacher--reference mismatch at initialization and changes in student output distributions after training.
The diagnostic configurations and their relationship to the main experiments are specified below.
All before-and-after comparisons use the same recorded contexts and token positions within each teacher setting.

\subsection{Diagnostic Data and Checkpoints}
\label{app:diagnosis-data}

For agentic search, teacher--reference gap statistics use the full reference-rollout caches described in Appendix~\ref{app:impl-search}.
Distributional measurements use the first $1{,}000$ cached trajectories, with further position sampling for the logit-margin analysis as specified below.
We compare $\pref$ with the final checkpoints from the main experiments at matched reference-generated contexts.
The diagnostic trajectories are drawn from the caches used for offline optimization.

For the cross-domain gap comparison, mathematical-reasoning rollouts are generated from $\pref$ on DAPO-Math-17k, following the one-response-per-prompt procedure of \citet{wu2026lightning}.
The reference checkpoint and generation settings are described in Appendix~\ref{app:impl-math}.
Both domains use Qwen3-32B as the scoring teacher.

\subsection{Teacher--Reference Gaps}
\label{app:diagnosis-gap}

At initialization, the recorded-token log-probability gap is
$\Delta_t=\log\teacher(a_t\mid s_t)-\log\pref(a_t\mid s_t)$.
Negative values indicate that the teacher assigns lower probability to the recorded token than the reference policy.
We compute the mean gap and threshold frequencies with equal weight for every scored model-generated token.
Prompt tokens, retrieved passages, and environment-inserted content are excluded.
Threshold frequencies are computed directly from the recorded gaps.

Figure~\ref{fig:gap} uses nonuniform bins whose widths increase geometrically away from zero.
Each bar reports the percentage of generated tokens falling within its bin, without normalization by bin width.
Dashed lines indicate the token-weighted mean gap for each domain.

\subsection{Distributional Measurements}
\label{app:diagnosis-metrics}

\paragraph{Next-token entropy.}
We compute Shannon entropy over the full $151{,}936$-token vocabulary in float64.
At each evaluated position, we subtract the reference entropy from the trained student's entropy.
We report the mean difference over all model-generated positions in the first $1{,}000$ cached trajectories, comprising $489{,}826$ positions for MiroThinker-1.7-mini and $714{,}255$ for Qwen3-32B.
Each position receives equal weight, so longer trajectories contribute more to the aggregate.
Entropy differences are reported in nats.

\paragraph{KL divergence to the teacher.}
Our primary alignment metric is the change in reverse KL, $\KL(p_t\|q_t)$, relative to $\pref$.
We additionally compute forward KL, $\KL(q_t\|p_t)$.
Both are evaluated in float64 on coarsened vocabulary partitions at selected positions with cached teacher-head scores.

The initial partition contains the teacher's probability-ranked prefix reaching cumulative mass $0.99$, capped at $32$ tokens, together with reference-proposed and recorded tokens whose teacher scores are available.
These scores come from the cached teacher top-$256$ scoring pass.
Tokens without available teacher scores are included in the residual bucket.
For reverse KL, the diagnostic implementation retains at most $32$ candidate cells and merges any omitted cells into the residual bucket, recomputing the corresponding teacher and student bucket masses.
For each KL direction, the partition and teacher probabilities are held fixed across checkpoints.

We subtract the reference KL at each position and average the differences with equal weight over the selected positions.
These measurements are reported in nats and describe alignment on the evaluated partitions.
With exact cell probabilities, coarsening provides a lower bound on full-vocabulary KL, but changes in coarsened KL do not necessarily bound changes in full-vocabulary KL.

\paragraph{Logit-margin ratio.}
We estimate relative logit scaling using ordinary least-squares regression of student log-probabilities on reference log-probabilities, with an intercept and equal token weights.
The regression uses tokens ranked $2$ through $1{,}000$ by the reference, excluding the recorded token.
Under an exact logit rescaling $z\mapsto\alpha z$, the fitted slope equals $\alpha$.

Positions are drawn from the first $1{,}000$ cached trajectories, sampling $300$ positions per token role and retaining all EOS and answer-closing positions.
The sampled positions are fixed across checkpoints.
We report median slopes and interquartile ranges separately by token role and by bins of the second-highest reference log-probability, with boundaries at $-13$, $-8$, and $-4$.
Slopes below one indicate contraction in the fitted relationship, without requiring the full student distribution to be an exact rescaling of the reference.

\begin{table}[t]
\centering
\footnotesize
\setlength{\tabcolsep}{3pt}
\begin{tabularx}{\columnwidth}{@{}lX@{}}
\toprule
\tableheader{Measurement} & \tableheader{Evaluation specification} \\
\midrule
Entropy
& Full vocabulary; all generated positions in the first
$1{,}000$ cached trajectories
($489{,}826$ / $714{,}255$ positions);
token-weighted mean change. \\

Teacher KL
& Selected positions in the same trajectories;
fixed candidate-and-bucket partitions for each KL direction;
token-weighted mean change. \\

Logit-margin ratio
& Positions sampled from the same trajectories:
$300$ per role, with all EOS and answer-closing positions retained;
regression over reference ranks $2$--$1{,}000$,
excluding the recorded token;
median and IQR within each role and reference-confidence bin. \\
\bottomrule
\end{tabularx}
\caption{Distributional diagnostic protocol.
Paired counts correspond to MiroThinker-1.7-mini and Qwen3-32B, respectively.
All checkpoints are evaluated at matched reference-generated contexts.
Entropy and KL changes are measured relative to $\pref$.}
\label{tab:diag-protocol}
\end{table}

\subsection{Diagnostic Results}
\label{app:diagnosis-results}

Table~\ref{tab:plane} reports changes in entropy and reverse KL relative to $\pref$, using the protocol described above.
Both metrics are expressed in nats.
Negative KL changes indicate improved teacher alignment on the evaluated partitions.

\begin{table}[t]
\centering
\footnotesize
\setlength{\tabcolsep}{4pt}
\begin{tabular}{@{}lcc@{}}
\toprule
\tableheader{Method}
& \tableheader{$\Delta H$}
& \tableheader{$\Delta\KL(p_t\|q_t)$} \\
\midrule
\multicolumn{3}{@{}l}{\tableheader{\textit{Teacher: MiroThinker-1.7-mini}}} \\
Online OPD
& $+0.81$ & $-0.06$ \\
\lopd{}
& $+1.20$ & $+0.18$ \\
\shortours{}
& $+0.11$ & $-0.17$ \\
\midrule
\multicolumn{3}{@{}l}{\tableheader{\textit{Teacher: Qwen3-32B}}} \\
Online OPD
& $+0.68$ & $-0.11$ \\
\lopd{}
& $+1.70$ & $+0.46$ \\
\shortours{}
& $+0.15$ & $-0.18$ \\
\bottomrule
\end{tabular}
\caption{Distributional changes relative to $\pref$, in nats.
Entropy is measured over all evaluated generated positions;
reverse KL is measured on fixed candidate-and-bucket partitions at selected positions.
Each method is compared with the reference at matched contexts.}
\label{tab:plane}
\end{table}

With both teachers, \lopd{} increases entropy and reverse KL, indicating less concentrated outputs and poorer alignment on the evaluated partitions.
In contrast, \shortours{} reduces reverse KL while producing substantially smaller entropy increases.
These measurements support the main-text finding that \shortours{} improves teacher alignment while mitigating the distributional flattening observed with \lopd{}.

\section{Qualitative Analysis}
\label{app:casestudy}

We examine two HotpotQA examples to illustrate failure behaviors observed in \lopd{} and contrast them with $\pref$ and \shortours{}.
The first concerns failure to produce a final answer, while the second concerns an incorrect answer presented as supported by retrieval despite no search being performed.

\paragraph{Evaluation and case selection.}
The examples come from the final evaluation with an eight-turn interaction budget, four responses per question, and three retrieved passages per search.
Both distillation methods use their final $150$-step checkpoints.
We display rollout~$0$ for each method and additionally report question-level F1 averaged over all four responses.
The cases are selected to illustrate specific behaviors rather than to estimate their prevalence or establish overall performance differences.
The number of searches in a displayed trajectory need not equal the interaction limit.
Quoted excerpts preserve the model output, with omissions marked by \texttt{[\ldots]}.

\begin{table}[t]
\centering
\footnotesize
\setlength{\tabcolsep}{4pt}
\begin{tabularx}{\columnwidth}{@{}Xrrr@{}}
\toprule
\tableheader{Method} & \tableheader{F1} & \tableheader{Searches} & \tableheader{Mean F1} \\
\midrule
\multicolumn{4}{@{}l}{\tableheader{\textit{MiroThinker-1.7-mini: question 222}}} \\
$\pref$       & $1.00$ & $2$ & $0.750$ \\
\lopd{}       & $0.00$ & $4$ & $0.750$ \\
\shortours{}  & $1.00$ & $4$ & $0.750$ \\
\midrule
\multicolumn{4}{@{}l}{\tableheader{\textit{Qwen3-32B: question 682}}} \\
$\pref$       & $0.67$ & $1$ & $0.750$ \\
\lopd{}       & $0.00$ & $0$ & $0.392$ \\
\shortours{}  & $1.00$ & $2$ & $0.683$ \\
\bottomrule
\end{tabularx}
\caption{Qualitative examples from HotpotQA.
F1 and search counts refer to the displayed rollout.
Mean F1 averages all four responses to the same question.
Scores are reported on the $[0,1]$ scale.}
\label{tab:case-arms}
\end{table}

\subsection{Failure to Produce a Final Answer}
\label{app:casestudy-miro}

The first question asks, ``Are both Dafeng District and Dazhou located in the same province?'', with the gold answer ``no''.
The reference policy searches for \texttt{Dafeng District province} and \texttt{Dazhou province}, retrieves evidence identifying Jiangsu and Sichuan, respectively, and answers correctly.
In contrast, \lopd{} performs four searches but produces no final answer.

The \lopd{} trajectory contains repeated self-correction before its first search:
\begin{quote}
\small
First, let's search for Dafeng District Dafenwise?
Wait my package: ``Dafeng District'' actually refers to Dafeng District of China.
Dafeng District is in the Chuanghai region?
Wait no, Dafeng District is in the province of Shubei?
Let me-check.
\texttt{[\ldots]}
\end{quote}

After receiving a passage identifying Dafeng District as part of Yancheng, Jiangsu, \lopd{} searches again for \texttt{Dafeng District Jiangsu}.
It subsequently produces the malformed search tag \texttt{<searchDazhou province</search>}, followed by a search for \texttt{dazhou mountain province}.
The latter retrieves information about a mountain in Guangdong rather than resolving the original comparison.
The displayed trajectory ends without an extractable final answer.

\shortours{} also performs four searches and returns ``NO'', but its reasoning incorrectly places Dafeng in Hebei and Dazhou in Anhui.
Its correct answer therefore does not establish that it accurately incorporated the retrieved evidence.
The observable distinction is narrower: \shortours{} produces an answer in this rollout, whereas \lopd{} does not.

All three methods obtain the same question-level F1 of $0.750$ over four responses.
This example illustrates a difference between individual trajectories, rather than a question-level performance gain.

\subsection{Answering Without Retrieval}
\label{app:casestudy-dense}

The second question asks, ``What show other than Hello Ross did Chelsea Handler appear on in January of 2016'', with the gold answer ``Chelsea Does''.
The reference policy performs one search and answers ``Chelsea'', receiving partial credit.
\shortours{} performs two searches and returns the complete answer.
\lopd{} performs no search and answers ``The Cold''.

Despite making no retrieval call, the \lopd{} response states:
\begin{quote}
\small
I should search for Chelsea Handler's appearances in January 2016.
Let me check the search results.

After the first search, the information mentions that she appeared on ``The Cold'' in January 2016.
That's the correct answer.
Since the initial search results confirm this, I can stop here and provide ``The Cold'' as the answer.
\end{quote}

The recorded trajectory contains neither a search call nor a retrieved observation supporting this assertion.
The model treats its own claim about nonexistent search results as confirmation and then emits the incorrect answer.
This provides a concrete example of answering without retrieval while presenting the answer as externally verified.

Both $\pref$ and \shortours{} retrieve passages from the Wikipedia pages \emph{Chelsea Handler} and \emph{Chelsea (TV series)}.
The example therefore does not establish that \shortours{} succeeds because it retrieves better sources.
It shows that the displayed \shortours{} response includes actual retrieval and the correct answer, whereas the \lopd{} response does neither.

Across all four responses, question-level F1 is $0.750$ for $\pref$, $0.392$ for \lopd{}, and $0.683$ for \shortours{}.
Thus, \shortours{} improves over \lopd{} on this question while remaining below the reference.
The aggregate comparisons in Table~\ref{tab:main}, rather than these selected trajectories, establish the overall performance differences.

\end{document}